\pdfoutput=1

\documentclass[11pt]{article}

\usepackage{naacl2021}

\usepackage{times}
\usepackage{latexsym}

\usepackage[T1]{fontenc}

\usepackage[utf8]{inputenc}

\usepackage{microtype}
\usepackage{pifont}
\usepackage{amssymb}
\usepackage{xcolor}
\usepackage{booktabs}
\usepackage{multicol}
\usepackage{multirow}
\usepackage{pgfplots}
\usepackage{enumitem}
\usepackage{amsmath}
\usepackage{graphicx}
\usepackage{algorithm}
\usepackage{algorithmic}
\usepackage{mathtools}
\usepackage{tikz}

\usepackage{color,soul}

\DeclareMathOperator*{\argmax}{argmax}
\DeclareMathOperator*{\topk}{arg^{k}max}
\DeclareMathSymbol{\shortminus}{\mathbin}{AMSa}{"39}

\newcommand{\cmark}{\ding{51}}%
\newcommand{\xmark}{\ding{55}}%
\newcommand{\Sref}[1]{\S\ref{#1}}

\newcommand{\Fref}[1]{Figure~\ref{#1}}

\newcommand{\Tref}[1]{Table~\ref{#1}}

\newlength\myindent
\pgfplotsset{compat=1.17}

\newcommand{\Hquad}{\hspace{0.0em}} 
\newcommand\mypar[1]{\noindent\textbf{#1}\Hquad}
\newcommand\aspace{\hspace{1.2em}}
\title{Searchable Hidden Intermediates for End-to-End Models of Decomposable Sequence Tasks}

\author{
Siddharth Dalmia \aspace Brian Yan \aspace Vikas Raunak \aspace Florian Metze \aspace Shinji Watanabe \\
Language Technologies Institute, Carnegie Mellon University, USA \\
\texttt{\{sdalmia,byan\}@cs.cmu.edu}
}

\begin{document}
\maketitle
\begin{abstract}
End-to-end approaches for sequence tasks are becoming increasingly popular. Yet for complex sequence tasks, like speech translation, systems that cascade several models trained on sub-tasks have shown to be superior, suggesting that the compositionality of cascaded systems simplifies learning and enables sophisticated search capabilities. In this work, we present an end-to-end framework that exploits compositionality to learn \textit{searchable} hidden representations at intermediate stages of a sequence model using decomposed sub-tasks. These hidden intermediates can be improved using beam search to enhance the overall performance and can also incorporate external models at intermediate stages of the network to re-score or adapt towards out-of-domain data. One instance of the proposed framework is a Multi-Decoder model for speech translation  that extracts the \textit{searchable hidden intermediates} from a speech recognition sub-task. The model demonstrates the aforementioned benefits and outperforms the previous state-of-the-art by around +6 and +3 BLEU on the two test sets of Fisher-CallHome and by around +3 and +4 BLEU on the English-German and English-French test sets of MuST-C.\footnote{All code and models are released as part of the ESPnet toolkit: \url{https://github.com/espnet/espnet}.}
\end{abstract}

\section{Introduction}
\label{sec:intro}

The principle of compositionality loosely states that a complex whole is composed of its parts and the rules by which those parts are combined \cite{lake2018generalization}. This principle is present in engineering, where task decomposition of a complex system is required to assess and optimize task allocations \cite{taskdecomp}, and in natural language, where paragraph coherence and discourse analysis rely on decomposition into sentences \cite{johnson1992, kuo1995} and sentence level semantics relies on decomposition into lexical units \cite{Liu2020}. 

Similarly, many sequence-to-sequence tasks that convert one sequence into another \cite{sutskever2014sequence} can be decomposed to simpler sequence sub-tasks in order to reduce the overall complexity. 
For example, speech translation systems, which seek to process speech in one language and output text in another language, can be naturally decomposed into the transcription of source language audio through automatic speech recognition (ASR) and translation into the target language through machine translation (MT). Such cascaded approaches have been widely used to build practical systems for a variety of sequence tasks like hybrid ASR \cite{hinton2012deep}, phrase-based MT \cite{koehn-etal-2007-moses}, and cascaded ASR-MT systems for speech translation (ST)~\cite{Pham2019TheI2}.

End-to-end sequence models like encoder-decoder models \cite{bahdanau2014neural, vaswani2017attention}, are attractive in part due to their simplistic design and the reduced need for hand-crafted features. However, studies have shown mixed results compared to cascaded models particularly for complex sequence tasks like speech translation \cite{inaguma-etal-2020-espnet-st} and spoken language understanding \cite{coucke2018snips}. Although direct target sequence prediction avoids the issue of error propagation from one system to another in cascaded approaches \cite{tzoukermann-miller-2018-evaluating-error}, there are many attractive properties of cascaded systems, missing in end-to-end approaches, that are useful in complex sequence tasks. 

In particular, we are interested in (1) the strong search capabilities of the cascaded systems that compose the final task output from individual system predictions \cite{mohri2002weighted, kumar2006weighted, beck-etal-2019-neural-lattice}, (2) the ability to incorporate external models to re-score each individual system \cite{och2002discriminative, huang2007forest}, (3) the ability to easily adapt individual components towards out-of-domain data \cite{koehn2007experiments, peddinti2015jhu}, and finally (4) the ability to monitor performance of the individual systems towards the decomposed sub-task \cite{tillmann2003word, meyer2016performance}.

In this paper, we seek to incorporate these properties of cascaded systems into end-to-end sequence models. We first propose a generic framework to learn \textit{searchable hidden intermediates} using an auto-regressive encoder-decoder model for any decomposable sequence task (\Sref{sec:methods}). We then apply this approach to speech translation, where the intermediate stage is the output of ASR, by passing continuous hidden representations of discrete transcript sequences from the ASR sub-net decoder to the MT sub-net encoder. By doing so, we gain the ability to use beam search with optional external model re-scoring on the hidden intermediates, while maintaining end-to-end differentiability. Next, we suggest mitigation strategies for the error propagation issues inherited from decomposition. 

We show the efficacy of \textit{searchable intermediate representations} in our proposed model, called the Multi-Decoder, on speech translation with a 5.4 and 2.8 BLEU score improvement over the previous state-of-the-arts for Fisher and CallHome test sets respectively (\Sref{sec:results}). We extend these improvements by an average of 0.5 BLEU score through the aforementioned benefit of re-scoring the intermediate search with external models trained on the same dataset. We also show a method for monitoring sub-net performance using oracle intermediates that are void of search errors (\Sref{sec:benefits}). Finally, we show how these models can adapt to out-of-domain speech translation datasets, how our approach can be generalized to other sequence tasks like speech recognition, and how the benefits of decomposition persist even for larger corpora like MuST-C (\Sref{sec:gen}).
\section{Background and Motivation}
\label{sec:background}

\begin{figure*}
\centering
\includegraphics[width=\linewidth]{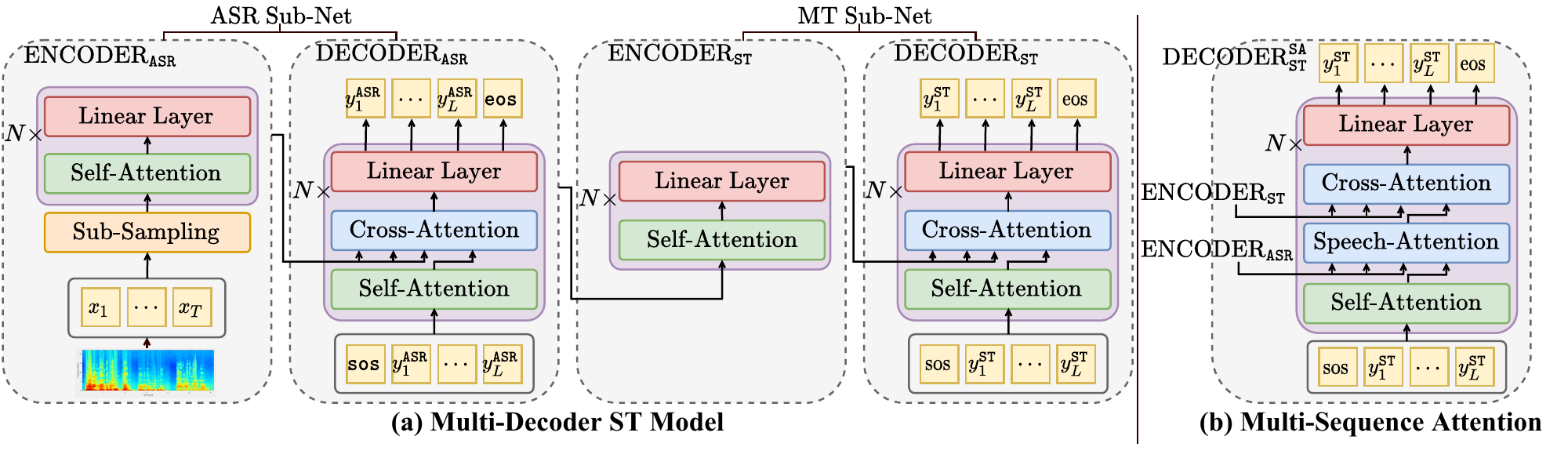}
\caption{The left side present the schematics and the information flow of our proposed framework applied to ST, in a model we call the Multi-Decoder. Our model decomposes ST into ASR and MT sub-nets, each of which consist of an encoder and decoder. The right side displays a Multi-Sequence Attention variant of the $\textsc{decoder}_{\textsc{st}}$ that is conditioned on both speech information via the $\textsc{encoder}_{\textsc{asr}}$ and transcription information via the $\textsc{encoder}_{\textsc{st}}$.}
\label{fig:model}
\end{figure*}

\subsection{Compositionality in Sequences Models}
The probabilistic space of a sequence is combinatorial in nature, such that a sentence of $L$ words from a fixed vocabulary $\mathcal{V}$ would have an output space $\mathcal{S}$ of size $|\mathcal{V}|^L$. In order to deal with this combinatorial output space, an output sentence is decomposed into labeled target tokens, $\mathbf{y} = (y_1, y_2,\ldots,y_L)$, where $y_l \in \mathcal{V}$. 
\begin{align*}
P(\mathbf{y} \mid \mathbf{x}) = \prod_{i=1}^{L} P(y_i \mid \mathbf{x}, y_{1:i\shortminus1})
\end{align*}
An auto-regressive encoder-decoder model uses the above probabilistic decomposition in sequence-to-sequence tasks to learn next word prediction,
which outputs a distribution over the next target token $y_l$ given the previous tokens $y_{1:l\shortminus1}$ and the input sequence $\mathbf{x} = (\mathbf{x}_1,\mathbf{x}_t,\ldots,\mathbf{x}_T)$, where $T$ is the input sequence length. In the next sub-section we detail the training and inference of these models.

\subsection{Auto-regressive Encoder-Decoder Models}
\label{sec:autoregressive}
\mypar{Training:} In an auto-regressive encoder-decoder model, the \textsc{Encoder} maps the input sequence $\mathbf{x}$ to a sequence of continuous hidden representations $\mathbf{h}^E = (\mathbf{h}_1^E,\mathbf{h}_t^E,\ldots,\mathbf{h}_T^E)$, where $\mathbf{h}_t^E \in \mathbb{R}^{d}$. The \textsc{Decoder} then auto-regressively maps $\mathbf{h}^E$ and the preceding ground-truth output tokens, $\hat{y}_{1:l\shortminus1}$, to $\mathbf{h}_l^D$, where $\mathbf{h}_l^D \in \mathbb{R}^{d}$. The sequence of decoder hidden representations form $\mathbf{h}^D = (\mathbf{h}_1^D,\mathbf{h}_l^D,\ldots,\mathbf{h}_L^D)$ and the likelihood of each output token ${y}_l$ is given by \textsc{SoftmaxOut}, which denotes an affine projection of $\mathbf{h}_l^D$ to $\mathcal{V}$ followed by a softmax function.
\begin{align}
\mathbf{h}^E &= \textsc{Encoder}(\mathbf{x}) \nonumber \\
\mathbf{\hat{h}}_l^D &= \textsc{Decoder}(\mathbf{h}^E, \hat{y}_{1:l\shortminus1}) \label{dec_eq1}\\
P(y_l \mid \hat{y}_{1:l\shortminus1}, \mathbf{h}^E) &= \textsc{SoftmaxOut}(\mathbf{\hat{h}}_l^D) \label{dec_softmax}
\end{align}
During training, the \textsc{Decoder} performs token classification for next word prediction by considering only the ground truth sequences for previous tokens $\mathbf{\hat{y}}$. We refer to this $\mathbf{\hat{h}}^D$ as \textit{oracle} decoder representations, which will be discussed later.

\mypar{Inference:} During inference, we can maximize the likelihood of the entire sequence from the output space $\mathcal{S}$ by composing the conditional probabilities of each step for the $L$ tokens in the sequence.
\begin{align}
\mathbf{h}_l^{D} &= \textsc{Decoder}(\mathbf{h}^E, y_{1:l\shortminus1}) \label{dec_eq2}\\
P(y_l \mid \mathbf{x}, y_{1:l\shortminus1}) &= \textsc{SoftmaxOut}(\mathbf{h}_l^{D}) \nonumber \\
\mathbf{\tilde{y}} = &\argmax_{\mathbf{y} \in \mathcal{S}} \prod_{i=1}^{L} P(y_i \mid \mathbf{x}, y_{1:i\shortminus1}) \label{inf_search}
\end{align}
This is an intractable search problem and it can be approximated by either greedily choosing $\argmax$ at each step or using a search algorithm like beam search to approximate $\mathbf{\tilde{y}}$. Beam search \cite{reddy1988foundations} generates candidates at each step and prunes the search space to a tractable beam size of $B$ most likely sequences. As $B \rightarrow \infty$, the beam search result would be equivalent to equation \ref{inf_search}. 
\begin{align*}
\textsc{GreedySearch} &\coloneqq \argmax_{y_l} P(y_l \mid \mathbf{x}, y_{1:l\shortminus1}) \\
\textsc{BeamSearch} &\coloneqq \textsc{Beam}(P(y_l \mid \mathbf{x}, y_{1:l\shortminus1}))
\end{align*}
In approximate search for auto-regressive models, like beam search, the \textsc{Decoder} receives alternate candidates of previous tokens to find candidates with a higher likelihood as an overall sequence. This also allows for the use of external models like Language Models (LM) or Connectionist Temporal Classification Models (CTC) for re-scoring candidates \cite{hori2017advances}.

\section{Proposed Framework}
\label{sec:methods}
In this section, we present a general framework to exploit natural decompositions in sequence tasks which seek to predict some output $\mathcal{C}$ from an input sequence $\mathcal{A}$. If there is an intermediate sequence $\mathcal{B}$ for which $\mathcal{A} \rightarrow \mathcal{B}$ sequence transduction followed by $\mathcal{B} \rightarrow \mathcal{C}$ prediction achieves the original task, then the original $\mathcal{A} \rightarrow \mathcal{C}$ task is decomposable. 

In other words, if we can learn $P(\mathcal{B} \mid \mathcal{A})$ then we can learn the overall task of $P(\mathcal{C} \mid \mathcal{A})$ through $\max_{\mathcal{B}}(P(\mathcal{C} \mid \mathcal{A}, ~\mathcal{B}) P(\mathcal{B} \mid \mathcal{A}))$, approximated using Viterbi search. We define a first encoder-decoder $\textsc{Sub}_{\mathcal{A} \to \mathcal{B}}\textsc{Net}$ to map an input sequence $\mathcal{A}$ to a sequence of decoder hidden states, $\mathbf{h}^{D_\mathcal{B}}$. Then we define a subsequent $\textsc{Sub}_{\mathcal{B} \to \mathcal{C}}\textsc{Net}$ to map $\mathbf{h}^{D_\mathcal{B}}$ to the final probabilistic output space of $\mathcal{C}$. Therefore, we call $\mathbf{h}^{D_\mathcal{B}}$ \textit{hidden intermediates}. The following equations shows the two sub-networks of our framework, $\textsc{Sub}_{\mathcal{A} \to \mathcal{B}}\textsc{Net}$ and $\textsc{Sub}_{\mathcal{B} \to \mathcal{C}}\textsc{Net}$, which can be trained end-to-end while also exploiting compositionality in sequence tasks.~\footnote{Note that this framework does not use locally-normalized softmax distributions but rather the hidden representations, thereby avoiding label bias issues when combining multiple sub-systems \cite{bottou1997global, wiseman2016sequence}.}

\paragraph{$\textsc{Sub}_{\mathcal{A} \to \mathcal{B}}\textsc{Net}$:}
\begin{align}
\mathbf{h}^E &= \textsc{Encoder}_\mathcal{A}(\mathcal{A}) \nonumber \\
\mathbf{\hat{h}}_l^{D_\mathcal{B}} &= \textsc{Decoder}_\mathcal{B}(\mathbf{h}^E, \mathbf{\hat{y}}^{\mathcal{B}}_{1:l\shortminus1}) \nonumber \\
P(y^{\mathcal{B}}_{l} \mid \mathbf{\hat{y}}^{\mathcal{B}}_{1:l\shortminus1}, \mathbf{h}^E) &= \textsc{SoftmaxOut}(\mathbf{\hat{h}}_l^{D_\mathcal{B}}) \label{intermediate_eq}
\end{align}
\paragraph{$\textsc{Sub}_{\mathcal{B} \to \mathcal{C}}\textsc{Net}$:}
\begin{equation}
P(\mathcal{C} \mid \mathbf{\hat{h}}_l^{D_\mathcal{B}}) = \textsc{Sub}_{\mathcal{B} \to \mathcal{C}}\textsc{Net}(\mathbf{\hat{h}}_l^{D_\mathcal{B}}) \label{generic_final}
\end{equation}
\noindent Note that the final prediction, given by equation \ref{generic_final}, does not need to be a sequence and can be a categorical class like in spoken language understanding tasks. Next we will show how the \textit{hidden intermediates} become \textit{searchable} during inference.

\subsection{Searchable Hidden Intermediates}

As stated in section \Sref{sec:autoregressive}, approximate search algorithms maximize the likelihood, $P(\mathbf{y}\mid\mathbf{x})$, of the entire sequence by considering different candidates  $y_l$ at each step. Candidate-based search, particularly in auto-regressive encoder-decoder models, also affects the decoder hidden representation, $\mathbf{h}^{D}$, as these are directly dependent on the previous candidate (refer to equations \ref{dec_eq1} and \ref{dec_eq2}). This implies that by searching for better approximations of the previous predicted tokens, $\mathbf{y}_{l \shortminus 1} = (\mathbf{y}_{\textsc{beam}})_{l \shortminus 1}$, we also improve the decoder hidden representations for the next token, $\mathbf{h}^{D}_l = (\mathbf{h}_{\textsc{beam}}^{D})_{l}$. As $\mathbf{y}_{\textsc{beam}} \to \mathbf{\hat{y}}$, the decoder hidden representations tend to the \textit{oracle} decoder representations that have only errors from next word prediction, $\mathbf{h}^{D}_{\textsc{beam}} \to \mathbf{\hat{h}}^D$. A perfect search is analogous to choosing the ground truth $\hat{y}$ at each step, which would yield $\mathbf{\hat{h}}^{D}$. 

We apply this beam search of hidden intermediates, thereby approximating $\mathbf{\hat{h}}^{D_\mathcal{B}}$ with $\mathbf{h}_{\textsc{beam}}^{D_\mathcal{B}}$. This process is illustrated in algorithm \ref{alg:asr_beam}, which shows beam search for $\mathbf{h}_\textsc{beam}^{D_\mathcal{B}}$ that are subsequently passed to the $\textsc{Sub}_{\mathcal{B} \to \mathcal{C}}\textsc{Net}$.\footnote{The algorithm shown only considers a single top approximation of the search; however, with added time-complexity, the final task prediction improves with the n-best $\mathbf{h}^{D_\mathcal{B}}_{\textsc{beam}}$ for selecting the best resultant $\mathcal{C}$.}
In line \ref{op_ext}, we show how an external model like an LM or a CTC model can be used to generate an alternate sequence likelihood, $P_\textsc{ext}(\mathbf{y}^{\mathcal{B}}_l)$, which can be combined with the $\textsc{Sub}_{\mathcal{A} \to \mathcal{B}}\textsc{Net}$ likelihood, $P_\mathcal{B}(\mathbf{y}^{\mathcal{B}}_l \mid \mathbf{x})$ , with a tunable parameter $\lambda$. 

\begin{algorithm}[h]
\caption{Beam Search for Hidden Intermediates: We perform beam search to approximate the most likely sequence for the sub-task $\mathcal{A} \to \mathcal{B}$, $\mathbf{y}^{\mathcal{B}}_{\textsc{beam}}$, while collecting the corresponding $\textsc{Decoder}_{\mathcal{B}}$ hidden representations, $\mathbf{h}^{D_\mathcal{B}}_{\textsc{beam}}$. The output $\mathbf{h}^{D_\mathcal{B}}_{\textsc{beam}}$, is passed to the final sub-network to predict final output $\mathcal{C}$ and $\mathbf{y}^{\mathcal{B}}_{\textsc{beam}}$ is used for monitoring performance on predicting $\mathcal{B}$.}
\label{alg:asr_beam}
\begin{algorithmic}[1]
\STATE \textbf{Initialize:} $\textsc{beam} \gets$ \{sos\}; k $\gets$ beam size;
\STATE $\mathbf{h}^{E_{\textsc{A}}}\gets \textsc{Encoder}_{\mathcal{A}}(\mathbf{x})$
\FOR{$l$=$1$ \textbf{to} $\max_{\textsc{steps}}$}
    \FOR{$\mathbf{y}^{\mathcal{B}}_{l\shortminus1} \in \textsc{beam}$}
        \STATE $\mathbf{h}^{D_\mathcal{B}}_{l} \gets \textsc{Decoder}_{\mathcal{B}}(\mathbf{h}^{E_{\mathcal{A}}}, \mathbf{y}^{\mathcal{B}}_{l\shortminus1})$
        \FOR{$\mathbf{y}^{\mathcal{B}}_{l} \in \mathbf{y}^{\mathcal{B}}_{l\shortminus1} + \{\mathcal{V}\}$}
            \STATE $s_{l} \gets P_{\mathcal{A} \to \mathcal{B}}(\mathbf{y}^{\mathcal{B}}_l \mid \mathbf{x})^{1 \shortminus \lambda} P_\textsc{ext}(\mathbf{y}^{\mathcal{B}}_l)^\lambda$ \label{op_ext}
            \STATE $\mathcal{H} \gets $ ($s_{l}$, $\mathbf{y}^{\mathcal{B}}_{l}$ , $\mathbf{h}^{D_\mathcal{B}}_{l}$)
        \ENDFOR
    \ENDFOR
    \STATE $\textsc{beam} \gets \topk(\mathcal{H})$
\ENDFOR
\STATE $(s^{\mathcal{B}}, \mathbf{y}^{\mathcal{B}}_{\textsc{beam}}, \mathbf{h}^{D_\mathcal{B}}_{\textsc{beam}}) \gets \argmax(\textsc{beam})$
\STATE \textbf{Return} $\mathbf{y}^{\mathcal{B}}_{\textsc{beam}} \rightarrow $  $\textsc{Sub}_{\mathcal{A} \to \mathcal{B}}\textsc{Net}$ Monitoring
\STATE \textbf{Return} $\mathbf{h}^{D_\mathcal{B}}_{\textsc{beam}} \rightarrow$ Final $\textsc{Sub}_{\mathcal{B} \to \mathcal{C}}\textsc{Net}$
\end{algorithmic}
\end{algorithm}
\noindent We can monitor the performance of the $\textsc{Sub}_{\mathcal{A} \to \mathcal{B}}\textsc{Net}$ by comparing the decoded intermediate sequence $\mathbf{y}^{\mathcal{B}}_{\textsc{beam}}$ to the ground truth $\mathbf{\hat{y}}^{\mathcal{B}}$. 
We can also monitor the $\textsc{Sub}_{\mathcal{B} \to \mathcal{C}}\textsc{Net}$ performance by using the aforementioned \textit{oracle} representations of the intermediates, $\mathbf{\hat{h}}^{D_\mathcal{B}}$, which can be obtained by feeding the ground truth $\mathbf{\hat{y}}^{\mathcal{B}}$ to $\textsc{Decoder}_\mathcal{B}$. By passing $\mathbf{\hat{h}}^{D_\mathcal{B}}$ to $\textsc{Sub}_{\mathcal{B} \to \mathcal{C}}\textsc{Net}$, we can observe its performance in a vacuum, i.e. void of search errors in the hidden intermediates.

\subsection{Multi-Decoder Model}
In order to show the applicability of our end-to-end framework we propose our Multi-Decoder model for speech translation. This model predicts a sequence of text translations $\mathbf{y}^{\textsc{st}}$ from an input sequence of speech $\mathbf{x}$ and uses a sequence of text transcriptions $\mathbf{y}^{\textsc{asr}}$ as an intermediate. In this case, the $\textsc{Sub}_{\mathcal{A} \to \mathcal{B}}\textsc{Net}$ in equation \ref{intermediate_eq} is specified as the ASR sub-net and the $\textsc{Sub}_{\mathcal{B} \to \mathcal{C}}\textsc{Net}$ in equation \ref{generic_final} is specified as the MT sub-net. Since the MT sub-net is also a sequence prediction task, both sub-nets are encoder-decoder models in our architecture~\cite{bahdanau2014neural, vaswani2017attention}. In Figure \ref{fig:model} we illustrate the schematics of our transformer based Multi-Decoder ST model which can also be summarized as follows:
\begin{align}
&\mathbf{h}^{E_{\textsc{asr}}} = \textsc{Encoder}_{\textsc{asr}}(\mathbf{x}) \label{eq_md_asr1}\\
&\mathbf{\hat{h}}_l^{D_\textsc{asr}} = \textsc{Decoder}_{\textsc{asr}}(\mathbf{h}^{E_{\textsc{asr}}}, \hat{y}^{\textsc{asr}}_{1:l\shortminus1}) \label{eq_md_asr2} \\
&\mathbf{h}^{E_\textsc{st}} = \textsc{Encoder}_{\textsc{st}}(\mathbf{\hat{h}}^{D_\textsc{asr}}) \label{eq_md1}\\
&\mathbf{\hat{h}}_l^{D_\textsc{st}} = \textsc{Decoder}_{\textsc{st}}(\mathbf{h}^{E_\textsc{st}}, \hat{y}^{\textsc{st}}_{1:l\shortminus1}) \label{eq_md2}
\end{align}
\noindent As we can see from Equations \ref{eq_md1} and \ref{eq_md2}, the MT sub-network attends only to the decoder representations, $\mathbf{\hat{h}}^{D_\textsc{asr}}$, of the ASR sub-network, which could lead to the error propagation issues from the ASR sub-network to the MT sub-network similar to the cascade systems, as mentioned in \Sref{sec:intro}. To alleviate this problem, we modify equation \ref{eq_md2} 
such that $\textsc{Decoder}_{\textsc{st}}$ attends to both $\mathbf{h}^{E_\textsc{st}}$ and $\mathbf{h}^{E_\textsc{asr}}$:
\begin{align}
\mathbf{\hat{h}}_l^{D^{\textsc{sa}}_\textsc{st}} = \textsc{Decoder}^{\textsc{sa}}_{\textsc{st}}(\mathbf{h}^{E_\textsc{st}}, \mathbf{h}^{E_\textsc{asr}}, \hat{y}^{\textsc{st}}_{1:l\shortminus1}) \label{eq_md_sa}
\end{align}
We use the multi-sequence cross-attention discussed by \citet{helcl-etal-2018-cuni}, shown on the right side of Figure \ref{fig:model}, to condition the final outputs generated by $\mathbf{\hat{h}}_l^{D_\textsc{st}}$ on both speech and transcript information in an attempt to allow our network to recover from intermediate mistakes during inference. We call this model the Multi-Decoder w/ Speech-Attention.

\section{Baseline Encoder-Decoder Model}
\label{sec:baseline}
For our baseline model, we use an end-to-end encoder-decoder (Enc-Dec) ST model with ASR joint training \cite{inaguma-etal-2020-espnet-st} as an auxiliarly loss to the speech encoder. In other words, the model consumes speech input using the $\textsc{Encoder}_\textsc{ASR}$, to produce $\mathbf{h}^{E_\textsc{asr}}$, which is used for cross-attention by $\textsc{Decoder}_\textsc{ASR}$ and the $\textsc{Decoder}_\textsc{ST}$. Using the decomposed ASR task as an auxiliary loss also helps the baseline Enc-Dec model and provide strong baseline performance, as we will see in Section \ref{sec:results}.

\section{Data and Experimental Setup}
\label{sec:data_prep}
\paragraph{Data:} We demonstrate the efficacy of our proposed approach on ST in the Fisher-CallHome corpus~\cite{postfisher} which contains 170 hours of Spanish conversational telephone speech, transcriptions, and English translations. All punctuations except apostrophes were removed and results are reported in terms of detokenized case-insensitive BLEU \cite{papineni-etal-2002-bleu, post-2018-call}. We compute BLEU using the 4 references in Fisher (dev, dev2, and test) and the single reference in CallHome (dev and test) \cite{postfisher, kumar2014, weiss2017}. We use a joint source and target vocabulary of 1K byte pair encoding (BPE) units \cite{kudo2018sentencepiece}.

We prepare the corpus using the ESPnet library and we follow the standard data preparation, where inputs are globally mean-variance normalized log-mel filterbank and pitch features from up-sampled 16kHz audio \cite{watanabe2018espnet}. We also apply speed perturbations of 0.9 and 1.1 and the SS SpecAugment policy \cite{specaug}.

\paragraph{Baseline Configuration:} 
All of our models are implemented using the ESPnet library and trained on 3 NVIDIA Titan 2080Ti GPUs for $\approx$12 hours. For the Baseline Enc-Dec baseline, discussed in \Sref{sec:baseline}, we use an $\textsc{Encoder}_{\textsc{asr}}$ consisting of a convolutional sub-sampling by a factor of 4 \cite{watanabe2018espnet} and 12 transformer encoder blocks with 2048 feed-forward dimension, 256 attention dimension, and 4 attention heads. The $\textsc{Decoder}_{\textsc{asr}}$ and $\textsc{Decoder}_{\textsc{st}}$ both consist of 6 transformer decoder blocks with the same configuration as $\textsc{Encoder}_{\textsc{asr}}$. There are 37.9M trainable parameters. We apply dropout of 0.1 for all components, detailed in the Appendix (\ref{sec:appendix}).

We train our models using an effective batch-size of 384 utterances and use the Adam optimizer \cite{kingma2014adam} with inverse square root decay learning rate schedule. We set learning rate to 12.5, warmup steps to 25K, and epochs to 50. We use joint training with hybrid CTC/attention ASR \cite{watanabe2017} by setting mtl-alpha to $0.3$ and asr-weight to $0.5$ as defined by \citet{watanabe2018espnet}. During inference, we perform beam search \cite{Seki2019VectorizedBS} on the ST sequences, using a beam size of 10, length penalty of 0.2, max length ratio of 0.3 \cite{watanabe2018espnet}. 

\paragraph{Multi-Decoder Configuration:}
For the Multi-Decoder ST model, discussed in \Sref{sec:methods}, we use the same transformer configuration as the baseline for the $\textsc{Encoder}_{\textsc{asr}}$, $\textsc{Decoder}_{\textsc{asr}}$, and $\textsc{Decoder}_{\textsc{st}}$. Additionally, the Multi-Decoder has an $\textsc{Encoder}_{\textsc{st}}$ consisting of 2 transformer encoder blocks with the same configuration as $\textsc{Encoder}_{\textsc{asr}}$, giving a total of 40.5M trainable parameters. The training configuration is also the same as for the baseline. For the Multi-Decoder w/ Speech-Attention model (42.1M trainable parameters), we increase the attention dropout of the ST decoder to 0.4 and dropout on all other components of the ST decoder to 0.2 while keeping dropout on the remaining components at 0.1. We verified that increasing the dropout does not help the vanilla multi-decoder ST model.

During inference, we perform beam search on both the ASR and ST output sequences, as discussed in \Sref{sec:methods}. The ST beam search is identical to that of the baseline. For the intermediate ASR beam search, we use a beam size of 16, length penalty of 0.2, max length ratio of 0.3. In some of our experiments, we also include fusion of a source language LM with a 0.2 weight and CTC with a 0.3 weight to re-score the intermediate ASR beam search~\cite{watanabe2017}. For the Speech-Attention variant, we increase LM weight to 0.4.

Note that the ST beam search configuration remains constant across our baseline and Multi-Decoder experiments as our focus is on improving overall performance through searchable intermediate representations. Thus, the various re-scoring techniques applied to the ASR beam search are options newly enabled by our proposed architecture and are not used in the ST beam search.

\begin{table*}[t]
  \centering
    \resizebox {\linewidth} {!} {
\begin{tabular}{llcccccc}
\toprule
& & Uses Speech  & \multicolumn{3}{c}{Fisher} & \multicolumn{2}{c}{CallHome}  \\
\cmidrule(r){4-6}\cmidrule(r){7-8}
Model Type & Model Name & Transcripts & dev($\uparrow$) & dev2($\uparrow$) & test($\uparrow$) & dev($\uparrow$) & test($\uparrow$) \\
\midrule
Cascade & \citet{inaguma-etal-2020-espnet-st} & \cmark & 41.5 & 43.5 & 42.2 & \textbf{19.6} & \textbf{19.8} \\
Cascade & ESPnet ASR+MT~\shortcite{watanabe2018espnet} & \cmark & \textbf{50.4} & \textbf{51.2} & \textbf{50.7} & \textbf{19.6} & 19.2 \\
\midrule
Enc-Dec & \citet{weiss2017}~$^\diamondsuit$ & \xmark & 46.5 & 47.3 & 47.3 & 16.4 & 16.6 \\
Enc-Dec & \citet{weiss2017}~$^\diamondsuit$ & \cmark & 48.3 & 49.1 & 48.7 & 16.8 & 17.4 \\
Enc-Dec & \citet{inaguma-etal-2020-espnet-st} & \cmark & 46.6 & 47.6 & 46.5 & 16.8 & 16.8 \\
Enc-Dec & \citet{guo2020recent} & \cmark & 48.7 & 49.6 & 47.0 & 18.5 & \textbf{18.6} \\
Enc-Dec & Our Implementation & \cmark & \textbf{49.6} & \textbf{50.9} & \textbf{49.5} & \textbf{19.1} & 18.2 \\
\midrule
Multi-Decoder & Our Proposed Model & \cmark & 52.7 & 53.3 & 52.6 & 20.5 & 20.1 \\ 
Multi-Decoder & \hspace{0.5em}+ASR Re-scoring & \cmark & 53.3 & 54.2 & 53.7 & 21.1 & 20.8 \\
Multi-Decoder & \hspace{0.5em}+Speech-Attention & \cmark & \textbf{54.6} & \textbf{54.6} & \textbf{54.1} & \textbf{21.7} & \textbf{21.4} \\
Multi-Decoder & \hspace{1em}+ASR Re-scoring & \cmark & \textbf{55.2} & \textbf{55.2} & \textbf{55.0} & \textbf{21.7} & \textbf{21.5} \\
\bottomrule
\end{tabular}
}
    \caption{Results presenting the overall performance (BLEU) of our proposed multi-decoder model. Cascade and Enc-Dec results from previous papers and our own implementation of the Enc-Dec are shown for comparison. The best performing models are \textbf{highlighted}. $^{\diamondsuit}$Implemented with LSTM, while all others are Transformer-based.}
    \label{tab:main_results}
\end{table*}
\section{Results}
\label{sec:results}

\Tref{tab:main_results} presents the overall ST performance (BLEU) of our proposed Multi-Decoder model. Our model improves by +2.9/+0.3 (Fisher/CallHome) over the best cascaded baseline and by +5.6/+1.5 BLEU over the best published end-to-end baselines. With Speech-Attention, our model improves by +3.4/+1.6 BLEU over the cascaded baselines and +7.1/+2.8 BLEU over encoder-decoder baselines. Both the Multi-Decoder and Multi-Decoder w/ Speech-Attention on average are further improved by +0.9/+0.4 BLEU through ASR re-scoring.\footnote{We also evaluate our models using other MT metrics to supplement these results, as shown in the Appendix (\ref{sec:other_metrics}).}

\Tref{tab:main_results} also includes our implementation of the Baseline Enc-Dec model discussed in \Sref{sec:baseline}. In this way, we are able to make a fair comparison with our framework as we control the model and inference configurations to be analagous. For instance, we keep the same search parameters for the final output in the baseline and the Multi-Decoder to demonstrate impact of the intermediate beam search.

\subsection{Benefits}

\label{sec:benefits}

\subsubsection{Sub-network performance monitoring}
\label{sec:pm}
An added benefit of our proposed approach over the Baseline Enc-Dec is the ability to monitor the individual performances of the ASR (\% WER) and MT (BLEU) sub-nets as shown in \Tref{tab:subnet_results}. The Multi-Decoder w/ Speech-Attention shows a greater MT sub-net performance than the Multi-Decoder as well as a slight improvement of the ASR sub-net, suggesting that ST can potentially help ASR. 
\begin{table}[t]
  \centering
    \resizebox {\linewidth} {!} {
\begin{tabular}{lc|cc}
\toprule
& Overall & Sub-Net & Sub-Net\\
Model & ST($\uparrow$) & ASR($\downarrow$) & MT($\uparrow$)\\
\midrule
Multi-Decoder & 52.7 & 22.6 & 64.9 \\
\hspace{0.5em} +Speech-Attention & 54.6 & 22.4 & 66.6 \\
\bottomrule
\end{tabular}
}
    \caption{Results presenting the overall ST performance (BLEU) of our Multi-Decoder models, along with their sub-net ASR (\% WER) and MT (BLEU) performances. All results are from the Fisher dev set.}
    \label{tab:subnet_results}
\end{table}

\subsubsection{Beam search for better intermediates}
\label{sec:beam}
The overall ST performance improves when a higher beam size is used in the intermediate ASR search, and this increase can be attributed to the improved ASR sub-net performance. \Fref{alg:asr_beam} shows this trend across ASR beam sizes of 1, 4, 8, 10, 16 while fixing the ST decoding beam size to 10. A beam size of 1, which is a greedy search, results in lower ASR sub-net and overall ST performances. As beam sizes become larger,  gains taper off as can be seen between beam sizes of 10 and 16.

\begin{figure}[t]
\resizebox {\linewidth} {!} {

\begin{tikzpicture}
	\begin{axis}[
		xlabel=ASR Beam Size,
		ylabel=ST BLEU Score ($\uparrow$),
		axis y line*=left,
		xtick=data,
	    every axis plot/.append style={thick},
		]
	\addplot[color=red,mark=triangle, mark options={scale=1.5}] coordinates {
	    (1,  51.55)
	    (4,  52.54)
	    (8,  52.63)
	    (10, 52.64)
	    (16, 52.66)
	}; \label{plot_one}
	\end{axis}
	
	\begin{axis}[
		axis y line*=right,
		axis x line=none,
		xtick=data,
		ylabel=ASR \% WER ($\downarrow$),
		legend cell align={right},
		legend style={at={(0.98,0.80)},anchor=north east},
	    every axis plot/.append style={thick},
		]
		\addlegendimage{empty legend}\addlegendentry{\hspace{-.6cm}\underline{Multi-Decoder}}
		\addlegendimage{/pgfplots/refstyle=plot_one}\addlegendentry{BLEU}
	\addplot[color=blue,mark=x, mark options={scale=1.5}] coordinates {
		(1,  23.7)
	    (4,  22.8)
	    (8,  22.7)
	    (10, 22.6)
	    (16, 22.6)
	}; \addlegendentry{\% WER}
	\end{axis}
\end{tikzpicture}
}
  \caption{Results studying the effect of the different ASR beam sizes in the intermediate representation search on the overall ST performance (BLEU) and the ASR sub-net performance (\%~WER) for our multi-decoder model. Beam of 1 is same as greedy search.}
  \label{fig:beam_exp}

\end{figure}
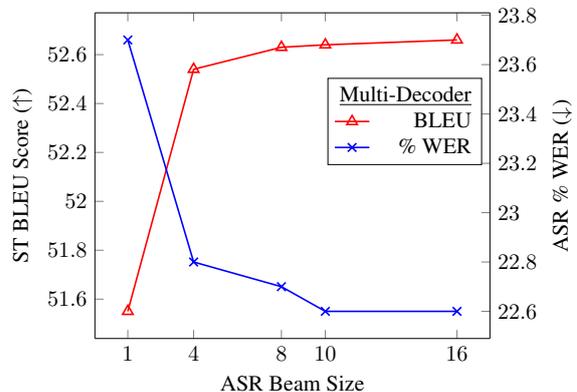

\subsubsection{External models for better search}
\label{sec:external}
External models like CTC acoustic models and language models are commonly used for re-scoring encoder-decoder models \cite{hori2017advances}, due to the difference in their modeling capabilities. CTC directly models transcripts while being conditionally independent on the other outputs given the input, and LMs predict the next token in a sequence.

Both variants of the Multi-Decoder improve due to improved ASR sub-net performance using external CTC and LM models for re-scoring, as shown in \Tref{tab:rescoring_results}. We use a recurrent neural network LM trained on the Fisher-CallHome Spanish transcripts with a dev perplexity of 18.8 and the CTC model from joint loss applied during training. Neither external model incorporates additional data. Although the impact of the LM-only re-scoring is not shown in the ASR \% WER, it reduces substitution and deletion rates in the ASR and this is observed to help the overall ST performance. 
\begin{table}[t]
  \centering
    \resizebox {\linewidth} {!} {
\begin{tabular}{lc|c}
\toprule
& Overall & Sub-Net \\
Model & ST($\uparrow$) & ASR($\downarrow$)\\
\midrule
Multi-Decoder & 52.7 & 22.6 \\
\hspace{0.5em}+ASR Re-scoring w/ LM & 53.2 & 22.6 \\
\hspace{0.5em}+ASR Re-scoring w/ CTC & 52.8 & 22.1 \\
\hspace{1.0em}+ASR Re-scoring w/ LM & \textbf{53.3} & \textbf{21.7} \\
\midrule
Multi-Decoder w/ Speech-Attn. & 54.6 & 22.4 \\
\hspace{0.5em}+ASR Re-scoring w/ LM & 55.1 & 22.4 \\
\hspace{0.5em}+ASR Re-scoring w/ CTC & 54.7 & 22.0 \\
\hspace{1.0em}+ASR Re-scoring w/ LM & \textbf{55.2} & \textbf{21.9} \\
\bottomrule
\end{tabular}
}
    \caption{Results presenting the overall ST performance (BLEU) and the sub-net ASR (\% WER) of our Multi-Decoder models with external CTC and LM re-scoring in the ASR intermediate representation search. All results are from the Fisher dev set.}
    \label{tab:rescoring_results}
\end{table}

\subsubsection{Error propagation avoidance}
\label{sec:error_prop}
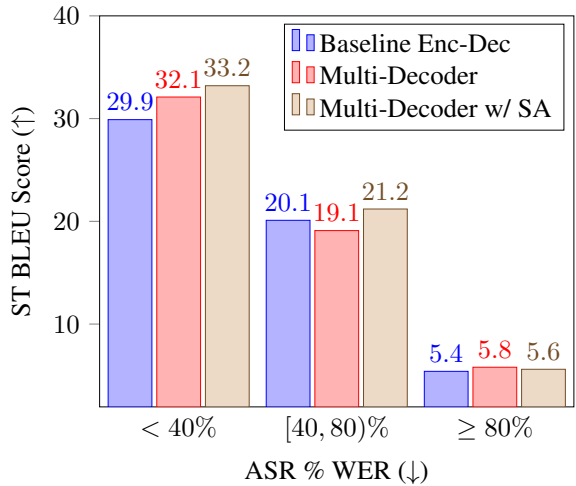
\begin{figure}[t]
\resizebox{\linewidth}{!}{
    \begin{tikzpicture}
        \begin{axis}[
            ybar,
            ymax=40,
            enlarge x limits=0.25,
            bar width=18pt,
            ylabel={ST BLEU Score ($\uparrow$)},
            xlabel={ASR \% WER ($\downarrow$)},
            symbolic x coords={$<40\%$,{$[40,80)\%$}, $\geq80\%$},
            xtick=data,
            nodes near coords,
            nodes near coords align={vertical},
            legend cell align={left},
            xlabel near ticks,
            ylabel near ticks,
            ytick align=outside,
            ytick pos=left,
            xtick align=inside,
            xtick style={draw=none},
            ]
            \addplot coordinates {($<40\%$,29.9) ({$[40,80)\%$},20.1) ($\geq80\%$,5.4)};
            \addplot coordinates {($<40\%$,32.1) ({$[40,80)\%$},19.1) ($\geq80\%$,5.8)};
            \addplot coordinates {($<40\%$,33.2) ({$[40,80)\%$},21.2) ($\geq80\%$,5.6)};
            \legend{Baseline Enc-Dec,Multi-Decoder,Multi-Decoder w/ SA}
        \end{axis}
    \end{tikzpicture}
    }
    \caption{Results comparing the ST performances (BLEU) of our Baseline Enc-Dec, Multi-Decoder, and Multi-Decoder w/ Speech-Attention across different ASR difficulties measured using \% WER on the Fisher dev set (1-ref). The buckets on the x-axis are determined using the utterance level \% WER using the Multi-Decoder ASR sub-net performance.}
    \label{fig:error_propagation}
\end{figure}

As discussed in \Sref{sec:methods}, our Multi-Decoder model inherits the error propagation issue as can be seen in \Fref{fig:error_propagation}. For the easiest bucket of utterances with $<40\%$ WER in Multi-Decoder's ASR sub-net, our model's ST performance, as measured by the corpus BLEU of the bucket, exceeds that of the Baseline Enc-Dec. The inverse is true for the more difficult bucket of $[40,80)\%$, showing that error propagation is limiting the performance of our model; however, we show that multi-sequence attention can alleviate this issue. For extremely difficult utterances in the $\geq80\%$ bucket, ST performance for all three approaches is suppressed. We also provide qualitative examples of error propagation avoidance in the Appendix (\ref{sec:qual_examples}).

\begin{table}[t]
  \centering
    \resizebox {\linewidth} {!} {
\begin{tabular}{lc|c}
\toprule
& Overall & Sub-Net\\
Model & ST($\uparrow$) & ASR($\downarrow$)\\
\midrule
\multicolumn{3}{l}{\textsc{\underline{In-domain ST Model}}} \\
Baseline~\cite{wang2020covost} & 12.0 & - \\
\hspace{0.5em}+ASR Pretrain~\cite{wang2020covost}~$^\diamondsuit$ & 23.0 & 16.0 \\
\midrule
\multicolumn{3}{l}{\textsc{\underline{Out-of-domain ST Model}}} \\
Multi-Decoder & 11.8 & 46.8 \\
\hspace{0.5em}+ASR Re-scoring w/ in-domain LM& 14.4 & \textbf{36.7} \\
Multi-Decoder w/ Speech-Attention & 12.6 & 46.5 \\
\hspace{0.5em}+ASR Re-scoring w/ in-domain LM & \textbf{15.0} & \textbf{36.7} \\
\bottomrule
\end{tabular}
}
    \caption{Results presenting the overall ST performance (BLEU) and the sub-net ASR (\% WER) of our Multi-Decoder models when tested on out-of-domain data. All models were trained on the Fisher-CallHome Es$\to$En corpus and tested on CoVost2 Es$\to$En corpus. $^\diamondsuit$Pretrained with 364 hours of in-domain ASR data.}
    \label{tab:robustness_results}
\end{table}

\subsection{Generalizability}
\label{sec:gen}
In this section, we discuss the generalizability of our framework towards out-of-domain data. We also extend our Multi-Decoder model to other sequence tasks like speech recognition. Finally, we apply our ST models to a larger corpus with more language pairs and a different domain of speech.

\subsubsection{Robustness through Decomposition}
\label{sec:robust_decomp}
Like cascaded systems, searchable intermediates provide our model adaptability in individual sub-systems towards out-of-domain data using external in-domain language model, thereby giving access to more in-domain data. Specifically for speech translation systems, this means we can use in-domain language models in both source and target languages. We test the robustness of our Multi-Decoder model trained on Fisher-CallHome conversational speech dataset on read speech CoVost-2 dataset \cite{wang2020covost}. In \Tref{tab:robustness_results} we show that re-scoring the ASR sub-net with an in-domain LM improves ASR with around 10.0\% lower WER, improving the overall ST performance by around +2.5 BLEU. Compared to an in-domain ST baseline \cite{Wang2020fairseqSF}, our out-of-domain Multi-Decoder with in-domain ASR re-scoring demonstrates the robustness of our approach.

\subsubsection{Decomposing Speech Transcripts}
\label{sec:asr_decomp}

We apply our generic framework to another decomposable sequence task, speech recognition, and show the results of various levels of decomposition in \Tref{tab:asr_results}. We show that with phoneme, character, or byte-pair encoding (BPE) sequences as intermediates, the Multi-Decoder presents strong results on both Fisher and CallHome test sets. We also observe that the BPE intermediates perform better than phoneme/character variants, which could be attributed to the reduced search capabilities of encoder-decoder models using beam search on longer sequences \cite{sountsov-sarawagi-2016-length} like in phoneme/character sequences.

\begin{table}[t]
  \centering
    \resizebox {\linewidth} {!} {
\begin{tabular}{lccc}
\toprule
& & Fisher & CallHome \\
Model & Intermediate & ASR($\downarrow$)& ASR($\downarrow$)\\
\midrule
Enc-Dec~$^\diamondsuit$ & - & 23.2 & 45.3 \\
\midrule
Multi-Decoder & Phoneme & 20.7 & 40.0 \\
Multi-Decoder & Character & 20.4 & 39.9 \\
Multi-Decoder & BPE100 & \textbf{19.7} & \textbf{38.9} \\
\bottomrule
\end{tabular}
}
    \caption{Results presenting the \% WER ASR performance when using the Multi-Decoder model on decomposed ASR task with phoneme, character, and BPE100 as intermediates. All results are from the Fisher-CallHome Spanish corpus.~$^\diamondsuit$\cite{weiss2017}}
    \label{tab:asr_results}
\end{table}

\subsubsection{Extending to MuST-C Language Pairs}
\label{sec:must_c}

In addition to our results using the 170 hours of the Spanish-English Fisher-CallHome corpus, in \Tref{tab:mustc_results} we show that our decompositional framework is also effective on larger ST corpora. In particular, we use 400 hours of English-German and 500 hours of English-French ST from the MuST-C corpus \cite{di2019must}.
Our Multi-Decoder model improves by +2.7 and +1.5 BLEU, in German and French respectively, over end-to-end baselines from prior works that do not use additional training data. We show that ASR re-scoring gives an additional +0.1 and +0.4 BLEU improvement.~\footnote{Details of the MuST-C data preparation and model parameters are detailed in Appendix (\ref{sec:mustc_data}).}

By extending our Multi-Decoder models to this MuST-C study, we show the generalizability of our approach across several dimensions of ST tasks. 
First, our approach consistently improves over baselines across multiple language-pairs. 
Second, our approach is robust to the distinct domains of telephone conversations from Fisher-CallHome and the TED-Talks from MuST-C. 
Finally, by scaling from 170 hours of Fisher-CallHome data to 500 hours of MuST-C data, we show that the benefits of decomposing sequence tasks with searchable hidden intermediates persist even with more data.

Furthermore, the performance of our Multi-Decoder models trained with only English-German or English-French ST data from MuST-C is comparable to other methods which incorporate larger external ASR and MT data in various ways. For instance, \citet{zheng2021fused} use 4700 hours of ASR data and 2M sentences of MT data for pretraining and multi-task learning. Similarly, \citet{bahar2021tight} use 2300 hours of ASR data and 27M sentences of MT data for pretraining. Our competitive performance without the use of any additional data highlights the data-efficient nature of our proposed end-to-end framework as opposed to the baseline encoder-decoder model, as pointed out by \citet{sperber2020speech}.

\begin{table}[t]
  \centering
    \resizebox {\linewidth} {!} {
\begin{tabular}{lc|c}
\toprule
& En$\rightarrow$De& En$\rightarrow$Fr \\
Model & ST($\uparrow$) & ST($\uparrow$)\\
\midrule
NeurST \cite{zhao2020neurst} & 22.9 & 33.3 \\
Fairseq S2T \cite{Wang2020fairseqSF} & 22.7 & 32.9 \\
ESPnet-ST \cite{inaguma-etal-2020-espnet-st} & 22.9 & 32.7 \\
Dual-Decoder \cite{le2020dualdecoder} & 23.6 & 33.5 \\
\midrule
Multi-Decoder w/ Speech-Attn. & 26.3 & 37.0 \\
\hspace{0.5em}+ASR Re-scoring & \textbf{26.4} & \textbf{37.4} \\
\bottomrule
\end{tabular}
}
    \caption{Results presenting the overall ST performance (BLEU) of our Multi-Decoder w/ Speech-Attention models with ASR re-scoring across two language-pairs, English-German (En$\rightarrow$De) and English-French (En$\rightarrow$Fr). All results are from the MuST-C tst-COMMON sets. All models use speech transcripts.}
    \label{tab:mustc_results}
\end{table}
\section{Discussion and Relation to Prior Work}
\label{sec:related}

\paragraph{Compositionality:}
A number of recent works have constructed composable neural network modules for tasks such as visual question answering \cite{andreas2016neural}, neural MT \cite{raunak2019compositionality}, and synthetic sequence-to-sequence tasks \cite{lake2019compositional}.
Modules that are first trained separately can subsequently be tightly integrated into a single end-to-end trainable model by passing differentiable soft decisions instead of discrete decisions in the intermediate stage \cite{bahar2021tight}. 
Further, even a single encoder-decoder model can be decomposed into modular components where the encoder and decoder modules have explicit functions \cite{sdalmia}.  

\paragraph{Joint Training with Sub-Tasks:}
End-to-end sequence models been shown to benefit from introducing joint training with sub-tasks as auxiliary loss functions for a variety of tasks like ASR \cite{kim2017joint}, ST \cite{salesky, liu2019synchronous, dong2020sdst, le2020dualdecoder}, SLU \cite{haghani2018audio}. They have been shown to induce structure \cite{belinkov2020linguistic} and improve the model performance \cite{toshniwal2017multitask}, but this joint training may reduce data efficiency if some sub-nets are not included in the final end-to-end model \cite{Sperber2019AttentionPassingMF, Wang_Wu_Liu_Yang_Zhou_2020}. Our framework avoids this sub-net waste at the cost of computational load during inference.

\paragraph{Speech Translation Decoders:}
Prior works have used ASR/MT decoding to improve the overall ST decoding through synchronous decoding \cite{liu2019synchronous}, dual decoding \cite{le2020dualdecoder}, and successive decoding \cite{dong2020sdst}. These works partially or fully decode ASR transcripts and use discrete intermediates to assist MT decoding. \citet{tu2017neural} and \citet{anastasopoulos-chiang-2018-tied} are closest to our multi-decoder ST model, however the benefits of our proposed framework are not entirely explored in these works.

\paragraph{Two-Pass Decoding:}
Two-pass decoding involves first predicting with one decoder and then re-evaluating with another decoder \cite{geng2018adaptive, sainath2019two, hu2020deliberation,rijhwani2020ocr}. The two decoders iterate on the same sequence, so there is no decomposition into sub-tasks in this method. On the other hand, our approach provides the subsequent decoder with a more structured representation than the input by decomposing the complexity of the overall task. Like two-pass decoding, our approach provides a sense of the future to the second decoder which allows it to correct mistakes from the previous first decoder. 

\paragraph{Auto-Regressive Decoding:} As auto-regressive decoders inherently learn a language model along with the task at hand, they tend to be domain specific \cite{samarakoon2018domain, muller-etal-2020-domain}. 
This can cause generalizability issues during inference \cite{murray-chiang-2018-correcting, yang2018breaking}, impacting the performance of both the task at hand and any downstream tasks. 
Our approach alleviates these problems through intermediate search, external models for intermediate re-scoring, and multi-sequence attention.

\section{Conclusion and Future Work}
\label{sec:conc}
We present searchable hidden intermediates for end-to-end models of decomposable sequence tasks. 
We show the efficacy of our Multi-Decoder model on the Fisher-CallHome Es$\rightarrow$En and MuST-C En$\rightarrow$De and En$\rightarrow$Fr speech translation corpora, achieving state-of-the-art results. 
We present various benefits in our framework, including sub-net performance monitoring, beam search for better hidden intermediates, external models for better search, and error propagation avoidance. 
Further, we demonstrate the flexibility of our framework towards out-of-domain tasks with the ability to adapt our sequence model at intermediate stages of decomposition. 
Finally, we show generalizability by training Multi-Decoder models for the speech recognition task at various levels of decomposition. 

We hope insights derived from our study stimulate research on tighter integrations between the benefits of cascaded and end-to-end sequence models. Exploiting searchable intermediates through beam search is just  the tip of the iceberg for search algorithms, as numerous approximate search techniques like diverse beam search \cite{diverse_beam_search} and best-first beam search \cite{meister2020best} have been recently proposed to improve diversity and approximation of the most-likely sequence. Incorporating differentiable lattice based search \cite{hannun2020differentiable} can also allow the subsequent sub-net to digest n-best representations.
\section{Acknowledgements}
This work started while Vikas Raunak was a student at CMU, he is now working as a Research Scientist at Microsoft. We thank Pengcheng Guo, Hirofumi Inaguma, Elizabeth Salesky, Maria Ryskina, Marta M\'endez Sim\'on and Vijay Viswanathan for their helpful discussion during the course of this project. We also thank the anonymous reviewers for their valuable feedback. This work used the Extreme Science and Engineering Discovery Environment (XSEDE) \cite{xsede}, which is supported by National Science Foundation grant number ACI-1548562. Specifically, it used the Bridges system \cite{bridges}, which is supported by NSF award number ACI-1445606, at the Pittsburgh Supercomputing Center (PSC). The work was supported in part by an AWS Machine Learning Research Award. This research was also supported in part the DARPA KAIROS program from the Air Force Research Laboratory under agreement number FA8750-19-2-0200. The U.S. Government is authorized to reproduce and distribute reprints for Governmental purposes not withstanding any copyright notation there on. The views and conclusions contained herein are those of the authors and should not be interpreted as necessarily representing the official policies or endorsements, either expressed or implied, of the Air Force Research Laboratory or the U.S. Government.
\bibliography{multi_decoder}

\begin{thebibliography}{79}
\expandafter\ifx\csname natexlab\endcsname\relax\def\natexlab#1{#1}\fi

\bibitem[{Anastasopoulos and Chiang(2018)}]{anastasopoulos-chiang-2018-tied}
Antonios Anastasopoulos and David Chiang. 2018.
\newblock \href {https://doi.org/10.18653/v1/N18-1008} {Tied multitask learning
  for neural speech translation}.
\newblock In \emph{Proceedings of the 2018 Conference of the North {A}merican
  Chapter of the Association for Computational Linguistics: Human Language
  Technologies, Volume 1 (Long Papers)}, pages 82--91, New Orleans, Louisiana.
  Association for Computational Linguistics.

\bibitem[{Andreas et~al.(2016)Andreas, Rohrbach, Darrell, and
  Klein}]{andreas2016neural}
Jacob Andreas, Marcus Rohrbach, Trevor Darrell, and Dan Klein. 2016.
\newblock \href {https://doi.org/10.1109/CVPR.2016.12} {Neural module
  networks}.
\newblock In \emph{2016 {IEEE} Conference on Computer Vision and Pattern
  Recognition, {CVPR} 2016, Las Vegas, NV, USA, June 27-30, 2016}, pages
  39--48. {IEEE} Computer Society.

\bibitem[{Bahar et~al.(2021)Bahar, Bieschke, Schl{\"u}ter, and
  Ney}]{bahar2021tight}
Parnia Bahar, Tobias Bieschke, Ralf Schl{\"u}ter, and Hermann Ney. 2021.
\newblock \href {https://doi.org/10.1109/SLT48900.2021.9383462} {Tight
  integrated end-to-end training for cascaded speech translation}.
\newblock In \emph{2021 IEEE Spoken Language Technology Workshop (SLT)}, pages
  950--957. IEEE.

\bibitem[{Bahdanau et~al.(2015)Bahdanau, Cho, and Bengio}]{bahdanau2014neural}
Dzmitry Bahdanau, Kyunghyun Cho, and Yoshua Bengio. 2015.
\newblock \href {http://arxiv.org/abs/1409.0473} {Neural machine translation by
  jointly learning to align and translate}.
\newblock In \emph{3rd International Conference on Learning Representations,
  {ICLR} 2015}.

\bibitem[{Banerjee and Lavie(2005)}]{banerjee2005meteor}
Satanjeev Banerjee and Alon Lavie. 2005.
\newblock \href {https://www.aclweb.org/anthology/W05-0909} {{METEOR}: An
  automatic metric for {MT} evaluation with improved correlation with human
  judgments}.
\newblock In \emph{Proceedings of the {ACL} Workshop on Intrinsic and Extrinsic
  Evaluation Measures for Machine Translation and/or Summarization}, pages
  65--72, Ann Arbor, Michigan. Association for Computational Linguistics.

\bibitem[{Beck et~al.(2019)Beck, Cohn, and
  Haffari}]{beck-etal-2019-neural-lattice}
Daniel Beck, Trevor Cohn, and Gholamreza Haffari. 2019.
\newblock \href {https://doi.org/10.18653/v1/D19-5304} {Neural speech
  translation using lattice transformations and graph networks}.
\newblock In \emph{Proceedings of the Thirteenth Workshop on Graph-Based
  Methods for Natural Language Processing (TextGraphs-13)}, pages 26--31, Hong
  Kong. Association for Computational Linguistics.

\bibitem[{Belinkov et~al.(2020)Belinkov, Durrani, Dalvi, Sajjad, and
  Glass}]{belinkov2020linguistic}
Yonatan Belinkov, Nadir Durrani, Fahim Dalvi, Hassan Sajjad, and James Glass.
  2020.
\newblock \href {https://dl.acm.org/doi/10.1162/coli_a_00367} {On the
  linguistic representational power of neural machine translation models}.
\newblock \emph{Computational Linguistics}, 46(1):1--52.

\bibitem[{Bottou et~al.(1997)Bottou, Bengio, and Le~Cun}]{bottou1997global}
L{\'e}on Bottou, Yoshua Bengio, and Yann Le~Cun. 1997.
\newblock \href {https://ieeexplore.ieee.org/document/609370} {Global training
  of document processing systems using graph transformer networks}.
\newblock In \emph{Proceedings of IEEE Computer Society Conference on Computer
  Vision and Pattern Recognition}, pages 489--494. IEEE.

\bibitem[{Coucke et~al.(2018)Coucke, Saade, Ball, Bluche, Caulier, Leroy,
  Doumouro, Gisselbrecht, Caltagirone, Lavril et~al.}]{coucke2018snips}
Alice Coucke, Alaa Saade, Adrien Ball, Th{\'e}odore Bluche, Alexandre Caulier,
  David Leroy, Cl{\'e}ment Doumouro, Thibault Gisselbrecht, Francesco
  Caltagirone, Thibaut Lavril, et~al. 2018.
\newblock \href {https://arxiv.org/pdf/1805.10190.pdf} {Snips voice platform:
  an embedded spoken language understanding system for private-by-design voice
  interfaces}.
\newblock In \emph{Privacy in Machine Learning and Artificial Intelligence
  workshop, ICML}.

\bibitem[{Dalmia et~al.(2019)Dalmia, Mohamed, Lewis, Metze, and
  Zettlemoyer}]{sdalmia}
Siddharth Dalmia, Abdelrahman Mohamed, Mike Lewis, Florian Metze, and Luke
  Zettlemoyer. 2019.
\newblock \href {https://arxiv.org/pdf/1911.03782.pdf} {{E}nforcing
  encoder-decoder modularity in sequence-to-sequence models}.
\newblock \emph{arXiv preprint arXiv:1911.03782}.

\bibitem[{Di~Gangi et~al.(2019)Di~Gangi, Cattoni, Bentivogli, Negri, and
  Turchi}]{di2019must}
Mattia~A. Di~Gangi, Roldano Cattoni, Luisa Bentivogli, Matteo Negri, and Marco
  Turchi. 2019.
\newblock \href {https://doi.org/10.18653/v1/N19-1202} {{M}u{ST}-{C}: a
  {M}ultilingual {S}peech {T}ranslation {C}orpus}.
\newblock In \emph{Proceedings of the 2019 Conference of the North {A}merican
  Chapter of the Association for Computational Linguistics: Human Language
  Technologies}, pages 2012--2017, Minneapolis, Minnesota. Association for
  Computational Linguistics.

\bibitem[{Dong et~al.(2020)Dong, Wang, Zhou, Xu, Xu, and Li}]{dong2020sdst}
Qianqian Dong, Mingxuan Wang, Hao Zhou, Shuang Xu, Bo~Xu, and Lei Li. 2020.
\newblock \href {https://doi.org/10.1609/aaai.v34i05.6360} {{SDST}: Successive
  decoding for speech-to-text translation}.
\newblock \emph{Proceedings of the Thirty-Fifth AAAI Conference on Artificial
  Intelligence}.

\bibitem[{Geng et~al.(2018)Geng, Feng, Qin, and Liu}]{geng2018adaptive}
Xinwei Geng, Xiaocheng Feng, Bing Qin, and Ting Liu. 2018.
\newblock \href {https://doi.org/10.18653/v1/D18-1048} {Adaptive multi-pass
  decoder for neural machine translation}.
\newblock In \emph{Proceedings of the 2018 Conference on Empirical Methods in
  Natural Language Processing}, pages 523--532, Brussels, Belgium. Association
  for Computational Linguistics.

\bibitem[{Guo et~al.(2021)Guo, Boyer, Chang, Hayashi, Higuchi, Inaguma, Kamo,
  Li, Garcia-Romero, Shi et~al.}]{guo2020recent}
Pengcheng Guo, Florian Boyer, Xuankai Chang, Tomoki Hayashi, Yosuke Higuchi,
  Hirofumi Inaguma, Naoyuki Kamo, Chenda Li, Daniel Garcia-Romero, Jiatong Shi,
  et~al. 2021.
\newblock \href {https://arxiv.org/abs/2010.13956} {Recent developments on
  {ESP}net toolkit boosted by conformer}.
\newblock In \emph{2021 IEEE international conference on acoustics, speech and
  signal processing (ICASSP)}. IEEE.

\bibitem[{Haghani et~al.(2018)Haghani, Narayanan, Bacchiani, Chuang, Gaur,
  Moreno, Prabhavalkar, Qu, and Waters}]{haghani2018audio}
Parisa Haghani, Arun Narayanan, Michiel Bacchiani, Galen Chuang, Neeraj Gaur,
  Pedro Moreno, Rohit Prabhavalkar, Zhongdi Qu, and Austin Waters. 2018.
\newblock \href {https://ieeexplore.ieee.org/document/8639043} {From audio to
  semantics: Approaches to end-to-end spoken language understanding}.
\newblock In \emph{2018 IEEE Spoken Language Technology Workshop (SLT)}, pages
  720--726. IEEE.

\bibitem[{Hannun et~al.(2020)Hannun, Pratap, Kahn, and
  Hsu}]{hannun2020differentiable}
Awni Hannun, Vineel Pratap, Jacob Kahn, and Wei-Ning Hsu. 2020.
\newblock \href {https://arxiv.org/pdf/2010.01003.pdf} {Differentiable weighted
  finite-state transducers}.
\newblock \emph{arXiv preprint arXiv:2010.01003}.

\bibitem[{Helcl et~al.(2018)Helcl, Libovick{\'y}, and
  Vari{\v{s}}}]{helcl-etal-2018-cuni}
Jind{\v{r}}ich Helcl, Jind{\v{r}}ich Libovick{\'y}, and Du{\v{s}}an
  Vari{\v{s}}. 2018.
\newblock \href {https://doi.org/10.18653/v1/W18-6441} {{CUNI} system for the
  {WMT}18 multimodal translation task}.
\newblock In \emph{Proceedings of the Third Conference on Machine Translation:
  Shared Task Papers}, pages 616--623, Belgium, Brussels. Association for
  Computational Linguistics.

\bibitem[{Hinton et~al.(2012)Hinton, Deng, Yu, Dahl, Mohamed, Jaitly, Senior,
  Vanhoucke, Nguyen, Sainath et~al.}]{hinton2012deep}
Geoffrey Hinton, Li~Deng, Dong Yu, George~E Dahl, Abdel-rahman Mohamed, Navdeep
  Jaitly, Andrew Senior, Vincent Vanhoucke, Patrick Nguyen, Tara~N Sainath,
  et~al. 2012.
\newblock \href {https://ieeexplore.ieee.org/document/6296526} {Deep neural
  networks for acoustic modeling in speech recognition: The shared views of
  four research groups}.
\newblock \emph{IEEE Signal processing magazine}, 29(6):82--97.

\bibitem[{Hori et~al.(2017)Hori, Watanabe, Zhang, and Chan}]{hori2017advances}
Takaaki Hori, Shinji Watanabe, Yu~Zhang, and William Chan. 2017.
\newblock \href {https://doi.org/10.21437/Interspeech.2017-1296} {Advances in
  joint {CTC}-attention based end-to-end speech recognition with a deep {CNN}
  encoder and {RNN-LM}}.
\newblock In \emph{Proc. Interspeech 2017}, pages 949--953.

\bibitem[{Hu et~al.(2020)Hu, Sainath, Pang, and
  Prabhavalkar}]{hu2020deliberation}
Ke~Hu, Tara~N Sainath, Ruoming Pang, and Rohit Prabhavalkar. 2020.
\newblock \href {https://ieeexplore.ieee.org/document/9053606} {Deliberation
  model based two-pass end-to-end speech recognition}.
\newblock In \emph{ICASSP 2020-2020 IEEE International Conference on Acoustics,
  Speech and Signal Processing (ICASSP)}, pages 7799--7803. IEEE.

\bibitem[{Huang and Chiang(2007)}]{huang2007forest}
Liang Huang and David Chiang. 2007.
\newblock \href {https://www.aclweb.org/anthology/P07-1019} {Forest
  {R}escoring: Faster decoding with integrated language models}.
\newblock In \emph{Proceedings of the 45th Annual Meeting of the Association of
  Computational Linguistics}, pages 144--151, Prague, Czech Republic.
  Association for Computational Linguistics.

\bibitem[{Inaguma et~al.(2020)Inaguma, Kiyono, Duh, Karita, Yalta, Hayashi, and
  Watanabe}]{inaguma-etal-2020-espnet-st}
Hirofumi Inaguma, Shun Kiyono, Kevin Duh, Shigeki Karita, Nelson Yalta, Tomoki
  Hayashi, and Shinji Watanabe. 2020.
\newblock \href {https://doi.org/10.18653/v1/2020.acl-demos.34} {{ESP}net-{ST}:
  All-in-one speech translation toolkit}.
\newblock In \emph{Proceedings of the 58th Annual Meeting of the Association
  for Computational Linguistics: System Demonstrations}, pages 302--311.
  Association for Computational Linguistics.

\bibitem[{Johnson(1992)}]{johnson1992}
Patricia Johnson. 1992.
\newblock \href {https://doi.org/10.1177/003368829202300201} {Cohesion and
  coherence in compositions in {M}alay and {E}nglish}.
\newblock \emph{RELC Journal}, 23(2):1--17.

\bibitem[{Kim et~al.(2017)Kim, Hori, and Watanabe}]{kim2017joint}
Suyoun Kim, Takaaki Hori, and Shinji Watanabe. 2017.
\newblock \href {https://ieeexplore.ieee.org/document/7953075} {Joint
  {CTC}-attention based end-to-end speech recognition using multi-task
  learning}.
\newblock In \emph{2017 IEEE international conference on acoustics, speech and
  signal processing (ICASSP)}, pages 4835--4839. IEEE.

\bibitem[{Kingma and Ba(2015)}]{kingma2014adam}
Diederik~P. Kingma and Jimmy Ba. 2015.
\newblock \href {http://arxiv.org/abs/1412.6980} {Adam: {A} method for
  stochastic optimization}.
\newblock In \emph{3rd International Conference on Learning Representations,
  {ICLR} 2015}.

\bibitem[{Koehn et~al.(2007)Koehn, Hoang, Birch, Callison-Burch, Federico,
  Bertoldi, Cowan, Shen, Moran, Zens, Dyer, Bojar, Constantin, and
  Herbst}]{koehn-etal-2007-moses}
Philipp Koehn, Hieu Hoang, Alexandra Birch, Chris Callison-Burch, Marcello
  Federico, Nicola Bertoldi, Brooke Cowan, Wade Shen, Christine Moran, Richard
  Zens, Chris Dyer, Ond{\v{r}}ej Bojar, Alexandra Constantin, and Evan Herbst.
  2007.
\newblock \href {https://www.aclweb.org/anthology/P07-2045} {{M}oses: Open
  source toolkit for statistical machine translation}.
\newblock In \emph{Proceedings of the 45th Annual Meeting of the Association
  for Computational Linguistics Companion Volume Proceedings of the Demo and
  Poster Sessions}, pages 177--180, Prague, Czech Republic. Association for
  Computational Linguistics.

\bibitem[{Koehn and Schroeder(2007)}]{koehn2007experiments}
Philipp Koehn and Josh Schroeder. 2007.
\newblock \href {https://www.aclweb.org/anthology/W07-0733/} {Experiments in
  domain adaptation for statistical machine translation}.
\newblock In \emph{Proceedings of the second workshop on statistical machine
  translation}, pages 224--227.

\bibitem[{Kudo and Richardson(2018)}]{kudo2018sentencepiece}
Taku Kudo and John Richardson. 2018.
\newblock \href {https://www.aclweb.org/anthology/D18-2012/} {{SentencePiece}:
  A simple and language independent subword tokenizer and detokenizer for
  neural text processing}.
\newblock In \emph{Proceedings of the 2018 Conference on Empirical Methods in
  Natural Language Processing: System Demonstrations}, pages 66--71.

\bibitem[{Kumar et~al.(2014)Kumar, Post, Povey, and Khudanpur}]{kumar2014}
Gaurav Kumar, Matt Post, Daniel Povey, and Sanjeev Khudanpur. 2014.
\newblock \href {https://doi.org/10.1109/ICASSP.2014.6854197} {Some insights
  from translating conversational telephone speech}.
\newblock In \emph{{IEEE} International Conference on Acoustics, Speech and
  Signal Processing, {ICASSP} 2014, Florence, Italy, May 4-9, 2014}, pages
  3231--3235. {IEEE}.

\bibitem[{Kumar et~al.(2006)Kumar, Deng, and Byrne}]{kumar2006weighted}
Shankar Kumar, Yonggang Deng, and William Byrne. 2006.
\newblock \href
  {https://www.cambridge.org/core/journals/natural-language-engineering/article/weighted-finite-state-transducer-translation-template-model-for-statistical-machine-translation/42D939EEE3C5C526F7562CE85C8EE14E}
  {A weighted finite state transducer translation template model for
  statistical machine translation}.
\newblock \emph{Natural Language Engineering}, 12(1):35--76.

\bibitem[{Kuo(1995)}]{kuo1995}
Chih-Hua Kuo. 1995.
\newblock \href {https://doi.org/10.1177/003368829502600103} {Cohesion and
  coherence in academic writing: From lexical choice to organization}.
\newblock \emph{RELC Journal}, 26(1):47--62.

\bibitem[{Lake(2019)}]{lake2019compositional}
Brenden~M Lake. 2019.
\newblock \href
  {https://papers.nips.cc/paper/2019/file/f4d0e2e7fc057a58f7ca4a391f01940a-Paper.pdf}
  {Compositional generalization through meta sequence-to-sequence learning}.
\newblock In \emph{Advances in Neural Information Processing Systems}, pages
  9791--9801.

\bibitem[{Lake and Baroni(2018)}]{lake2018generalization}
Brenden~M. Lake and Marco Baroni. 2018.
\newblock \href {http://proceedings.mlr.press/v80/lake18a.html} {Generalization
  without systematicity: On the compositional skills of sequence-to-sequence
  recurrent networks}.
\newblock In \emph{Proceedings of the 35th International Conference on Machine
  Learning, {ICML} 2018}, pages 2879--2888. {PMLR}.

\bibitem[{Le et~al.(2020)Le, Pino, Wang, Gu, Schwab, and
  Besacier}]{le2020dualdecoder}
Hang Le, Juan Pino, Changhan Wang, Jiatao Gu, Didier Schwab, and Laurent
  Besacier. 2020.
\newblock \href {https://www.aclweb.org/anthology/2020.coling-main.314.pdf}
  {Dual-decoder transformer for joint automatic speech recognition and
  multilingual speech translation}.
\newblock \emph{Proceedings of the 28th International Conference on
  Computational Linguistics}.

\bibitem[{Levis et~al.(1994)Levis, Moray, and Hu}]{taskdecomp}
Alexander~H. Levis, Neville Moray, and Baosheng Hu. 1994.
\newblock \href {https://doi.org/https://doi.org/10.1016/0005-1098(94)90025-6}
  {Task decomposition and allocation problems and discrete event systems}.
\newblock \emph{Automatica}, 30(2):203 -- 216.

\bibitem[{Liu et~al.(2020{\natexlab{a}})Liu, Zhang, Xiong, Zhou, He, Wu, Wang,
  and Zong}]{liu2019synchronous}
Yuchen Liu, Jiajun Zhang, Hao Xiong, Long Zhou, Zhongjun He, Hua Wu, Haifeng
  Wang, and Chengqing Zong. 2020{\natexlab{a}}.
\newblock \href {https://aaai.org/ojs/index.php/AAAI/article/view/6360}
  {Synchronous speech recognition and speech-to-text translation with
  interactive decoding}.
\newblock In \emph{The Thirty-Fourth {AAAI} Conference on Artificial
  Intelligence, {AAAI} 2020}, pages 8417--8424.

\bibitem[{Liu et~al.(2020{\natexlab{b}})Liu, Lin, and Sun}]{Liu2020}
Zhiyuan Liu, Yankai Lin, and Maosong Sun. 2020{\natexlab{b}}.
\newblock \href {https://doi.org/10.1007/978-981-15-5573-2_3}
  {\emph{Compositional Semantics}}, pages 43--57. Springer Singapore,
  Singapore.

\bibitem[{Meister et~al.(2020)Meister, Vieira, and Cotterell}]{meister2020best}
Clara Meister, Tim Vieira, and Ryan Cotterell. 2020.
\newblock \href {https://doi.org/10.1162/tacl_a_00346} {Best-first beam
  search}.
\newblock \emph{Transactions of the Association for Computational Linguistics},
  8:795--809.

\bibitem[{Meyer et~al.(2016)Meyer, Mallidi, Martinez, Pay{\'a}-Vay{\'a},
  Kayser, and Hermansky}]{meyer2016performance}
Bernd~T Meyer, Sri~Harish Mallidi, Angel Mario~Castro Martinez, Guillermo
  Pay{\'a}-Vay{\'a}, Hendrik Kayser, and Hynek Hermansky. 2016.
\newblock \href {https://ieeexplore.ieee.org/document/7846244} {Performance
  monitoring for automatic speech recognition in noisy multi-channel
  environments}.
\newblock In \emph{2016 IEEE Spoken Language Technology Workshop (SLT)}, pages
  50--56.

\bibitem[{Mohri et~al.(2002)Mohri, Pereira, and Riley}]{mohri2002weighted}
Mehryar Mohri, Fernando Pereira, and Michael Riley. 2002.
\newblock \href
  {https://www.sciencedirect.com/science/article/abs/pii/S0885230801901846}
  {Weighted finite-state transducers in speech recognition}.
\newblock \emph{Computer Speech \& Language}, 16(1):69--88.

\bibitem[{M{\"u}ller et~al.(2020)M{\"u}ller, Rios, and
  Sennrich}]{muller-etal-2020-domain}
Mathias M{\"u}ller, Annette Rios, and Rico Sennrich. 2020.
\newblock \href {https://www.aclweb.org/anthology/2020.amta-research.14}
  {Domain robustness in neural machine translation}.
\newblock In \emph{Proceedings of the 14th Conference of the Association for
  Machine Translation in the Americas}, pages 151--164, Virtual. Association
  for Machine Translation in the Americas.

\bibitem[{Murray and Chiang(2018)}]{murray-chiang-2018-correcting}
Kenton Murray and David Chiang. 2018.
\newblock \href {https://doi.org/10.18653/v1/W18-6322} {Correcting length bias
  in neural machine translation}.
\newblock In \emph{Proceedings of the Third Conference on Machine Translation:
  Research Papers}, pages 212--223, Brussels, Belgium.

\bibitem[{Nystrom et~al.(2015)Nystrom, Levine, Roskies, and Scott}]{bridges}
Nicholas~A. Nystrom, Michael~J. Levine, Ralph~Z. Roskies, and J.~Ray Scott.
  2015.
\newblock \href {https://doi.org/10.1145/2792745.2792775} {Bridges: A uniquely
  flexible {HPC} resource for new communities and data analytics}.
\newblock In \emph{Proceedings of the 2015 XSEDE Conference: Scientific
  Advancements Enabled by Enhanced Cyberinfrastructure}. Association for
  Computing Machinery.

\bibitem[{Och and Ney(2002)}]{och2002discriminative}
Franz~Josef Och and Hermann Ney. 2002.
\newblock \href {https://www.aclweb.org/anthology/P02-1038/} {Discriminative
  training and maximum entropy models for statistical machine translation}.
\newblock In \emph{Proceedings of the 40th Annual meeting of the Association
  for Computational Linguistics}, pages 295--302.

\bibitem[{Papineni et~al.(2002)Papineni, Roukos, Ward, and
  Zhu}]{papineni-etal-2002-bleu}
Kishore Papineni, Salim Roukos, Todd Ward, and Wei-Jing Zhu. 2002.
\newblock \href {https://doi.org/10.3115/1073083.1073135} {{BLEU}: a method for
  automatic evaluation of machine translation}.
\newblock In \emph{Proceedings of the 40th Annual Meeting of the Association
  for Computational Linguistics}, pages 311--318, Philadelphia, Pennsylvania,
  USA.

\bibitem[{Park et~al.(2019)Park, Chan, Zhang, Chiu, Zoph, Cubuk, and
  Le}]{specaug}
Daniel~S Park, William Chan, Yu~Zhang, Chung-Cheng Chiu, Barret Zoph, Ekin~D
  Cubuk, and Quoc~V Le. 2019.
\newblock \href
  {https://www.isca-speech.org/archive/Interspeech_2019/pdfs/2680.pdf}
  {{SpecAugment}: A simple data augmentation method for automatic speech
  recognition}.
\newblock \emph{Proc. Interspeech 2019}, pages 2613--2617.

\bibitem[{Peddinti et~al.(2015)Peddinti, Chen, Manohar, Ko, Povey, and
  Khudanpur}]{peddinti2015jhu}
Vijayaditya Peddinti, Guoguo Chen, Vimal Manohar, Tom Ko, Daniel Povey, and
  Sanjeev Khudanpur. 2015.
\newblock \href {https://ieeexplore.ieee.org/document/7404842} {{JHU ASpIRE}
  system: Robust {LVCSR} with {TDNNS}, {iVector} adaptation and {RNN-LMS}}.
\newblock In \emph{2015 IEEE Workshop on Automatic Speech Recognition and
  Understanding (ASRU)}, pages 539--546.

\bibitem[{Pham et~al.(2019)Pham, Nguyen, Ha, Hussain, Schneider, Niehues,
  St{\"u}ker, and Waibel}]{Pham2019TheI2}
N.~Pham, Thai-Son Nguyen, Thanh-Le Ha, J.~Hussain, Felix Schneider, J.~Niehues,
  Sebastian St{\"u}ker, and A.~Waibel. 2019.
\newblock \href {https://zenodo.org/record/3525564} {The {IWSLT} 2019 {KIT}
  speech translation system}.
\newblock In \emph{International Workshop on Spoken Language Translation
  (IWSLT)}.

\bibitem[{Post(2018)}]{post-2018-call}
Matt Post. 2018.
\newblock \href {https://www.aclweb.org/anthology/W18-6319} {A call for clarity
  in reporting {BLEU} scores}.
\newblock In \emph{Proceedings of the Third Conference on Machine Translation:
  Research Papers}, pages 186--191, Belgium, Brussels. Association for
  Computational Linguistics.

\bibitem[{Post et~al.(2013)Post, Kumar, Lopez, Karakos, Callison-Burch, and
  Khudanpur}]{postfisher}
Matt Post, Gaurav Kumar, Adam Lopez, Damianos Karakos, Chris Callison-Burch,
  and Sanjeev Khudanpur. 2013.
\newblock \href {https://catalog.ldc.upenn.edu/LDC2014T23} {Improved
  speech-to-text translation with the {Fisher} and {Callhome}
  {Spanish–English} speech translation corpus}.
\newblock In \emph{International Workshop on Spoken Language Translation (IWSLT
  2013)}.

\bibitem[{Raunak et~al.(2019)Raunak, Kumar, and
  Metze}]{raunak2019compositionality}
Vikas Raunak, Vaibhav Kumar, and Florian Metze. 2019.
\newblock \href {https://arxiv.org/abs/1911.01497} {On compositionality in
  neural machine translation}.
\newblock \emph{NeurIPS Workshop, Context and Compositionality in Biological
  and Artificial Neural Systems}.

\bibitem[{Reddy(1988)}]{reddy1988foundations}
Raj Reddy. 1988.
\newblock \href {http://www.aaai.org/ojs/index.php/aimagazine/article/view/950}
  {Foundations and grand challenges of artificial intelligence: {AAAI}
  presidential address}.
\newblock \emph{{AI} Mag.}, 9(4):9--21.

\bibitem[{Rijhwani et~al.(2020)Rijhwani, Anastasopoulos, and
  Neubig}]{rijhwani2020ocr}
Shruti Rijhwani, Antonios Anastasopoulos, and Graham Neubig. 2020.
\newblock \href {https://www.aclweb.org/anthology/2020.emnlp-main.478} {{OCR}
  post correction for endangered language texts}.
\newblock In \emph{Proceedings of the 2020 Conference on Empirical Methods in
  Natural Language Processing (EMNLP)}, pages 5931--5942, Online. Association
  for Computational Linguistics.

\bibitem[{Sainath et~al.(2019)Sainath, Pang, Rybach, He, Prabhavalkar, Li,
  Visontai, Liang, Strohman, Wu et~al.}]{sainath2019two}
Tara~N Sainath, Ruoming Pang, David Rybach, Yanzhang He, Rohit Prabhavalkar,
  Wei Li, Mirk{\'o} Visontai, Qiao Liang, Trevor Strohman, Yonghui Wu, et~al.
  2019.
\newblock \href
  {https://www.isca-speech.org/archive/Interspeech_2019/pdfs/1341.pdf}
  {Two-pass end-to-end speech recognition}.
\newblock \emph{Proc. Interspeech 2019}, pages 2773--2777.

\bibitem[{Salesky et~al.(2019)Salesky, Sperber, and Black}]{salesky}
Elizabeth Salesky, Matthias Sperber, and Alan~W Black. 2019.
\newblock \href {https://doi.org/10.18653/v1/P19-1179} {{Exploring
  Phoneme-Level Speech Representations for End-to-End Speech Translation}}.
\newblock In \emph{Proceedings of the 57th Annual Meeting of the Association
  for Computational Linguistics}, pages 1835--1841, Florence, Italy.
  Association for Computational Linguistics.

\bibitem[{Samarakoon et~al.(2018)Samarakoon, Mak, and
  Lam}]{samarakoon2018domain}
Lahiru Samarakoon, Brian Mak, and Albert~YS Lam. 2018.
\newblock \href {https://ieeexplore.ieee.org/abstract/document/8639506} {Domain
  adaptation of end-to-end speech recognition in low-resource settings}.
\newblock In \emph{2018 IEEE Spoken Language Technology Workshop (SLT)}, pages
  382--388. IEEE.

\bibitem[{Seki et~al.(2019)Seki, Hori, Watanabe, Moritz, and
  Roux}]{Seki2019VectorizedBS}
Hiroshi Seki, Takaaki Hori, Shinji Watanabe, Niko Moritz, and Jonathan~Le Roux.
  2019.
\newblock \href {https://doi.org/10.21437/Interspeech.2019-2860} {Vectorized
  beam search for {CTC}-attention-based speech recognition}.
\newblock In \emph{Proc. Interspeech 2019}, pages 3825--3829.

\bibitem[{Snover et~al.(2006)Snover, Dorr, Schwartz, Micciulla, and
  Makhoul}]{snover2006study}
Matthew Snover, Bonnie Dorr, Richard Schwartz, Linnea Micciulla, and John
  Makhoul. 2006.
\newblock \href {http://www.cs.umd.edu/~snover/pub/amta06/ter_amta.pdf} {A
  study of translation edit rate with targeted human annotation}.
\newblock In \emph{Proceedings of Association for Machine Translation in the
  Americas}.

\bibitem[{Sountsov and Sarawagi(2016)}]{sountsov-sarawagi-2016-length}
Pavel Sountsov and Sunita Sarawagi. 2016.
\newblock \href {https://doi.org/10.18653/v1/D16-1158} {Length bias in encoder
  decoder models and a case for global conditioning}.
\newblock In \emph{Proceedings of the 2016 Conference on Empirical Methods in
  Natural Language Processing}, pages 1516--1525, Austin, Texas. Association
  for Computational Linguistics.

\bibitem[{Sperber et~al.(2019)Sperber, Neubig, Niehues, and
  Waibel}]{Sperber2019AttentionPassingMF}
Matthias Sperber, Graham Neubig, Jan Niehues, and Alex Waibel. 2019.
\newblock \href {https://doi.org/10.1162/tacl_a_00270} {Attention-passing
  models for robust and data-efficient end-to-end speech translation}.
\newblock \emph{Transactions of the Association for Computational Linguistics},
  7:313--325.

\bibitem[{Sperber and Paulik(2020)}]{sperber2020speech}
Matthias Sperber and Matthias Paulik. 2020.
\newblock \href {https://www.aclweb.org/anthology/2020.acl-main.661.pdf}
  {Speech translation and the end-to-end promise: Taking stock of where we
  are}.
\newblock \emph{Proceedings of the 58th Annual Meeting of the Association for
  Computational Linguistics}.

\bibitem[{Sutskever et~al.(2014)Sutskever, Vinyals, and
  Le}]{sutskever2014sequence}
Ilya Sutskever, Oriol Vinyals, and Quoc~V Le. 2014.
\newblock \href
  {https://papers.nips.cc/paper/2014/file/a14ac55a4f27472c5d894ec1c3c743d2-Paper.pdf}
  {Sequence to sequence learning with neural networks}.
\newblock In \emph{Advances in neural information processing systems}, pages
  3104--3112.

\bibitem[{Tillmann and Ney(2003)}]{tillmann2003word}
Christoph Tillmann and Hermann Ney. 2003.
\newblock \href {https://www.aclweb.org/anthology/J03-1005.pdf} {Word
  reordering and a dynamic programming beam search algorithm for statistical
  machine translation}.
\newblock \emph{Computational linguistics}, 29(1):97--133.

\bibitem[{Toshniwal et~al.(2017)Toshniwal, Tang, Lu, and
  Livescu}]{toshniwal2017multitask}
Shubham Toshniwal, Hao Tang, Liang Lu, and Karen Livescu. 2017.
\newblock \href {https://doi.org/10.21437/Interspeech.2017-1118} {Multitask
  learning with low-level auxiliary tasks for encoder-decoder based speech
  recognition}.
\newblock In \emph{Proc. Interspeech 2017}, pages 3532--3536.

\bibitem[{Towns et~al.(2014)Towns, Cockerill, Dahan, Foster, Gaither, Grimshaw,
  Hazlewood, Lathrop, Lifka, Peterson et~al.}]{xsede}
John Towns, Timothy Cockerill, Maytal Dahan, Ian Foster, Kelly Gaither, Andrew
  Grimshaw, Victor Hazlewood, Scott Lathrop, Dave Lifka, Gregory~D Peterson,
  et~al. 2014.
\newblock \href {https://ieeexplore.ieee.org/document/6866038} {{XSEDE}:
  accelerating scientific discovery}.
\newblock \emph{Computing in science \& engineering}, 16(5):62--74.

\bibitem[{Tu et~al.(2017)Tu, Liu, Shang, Liu, and Li}]{tu2017neural}
Zhaopeng Tu, Yang Liu, Lifeng Shang, Xiaohua Liu, and Hang Li. 2017.
\newblock \href {http://aaai.org/ocs/index.php/AAAI/AAAI17/paper/view/14161}
  {Neural machine translation with reconstruction}.
\newblock In \emph{Proceedings of the Thirty-First {AAAI} Conference on
  Artificial Intelligence, 2017}, pages 3097--3103. {AAAI} Press.

\bibitem[{Tzoukermann and
  Miller(2018)}]{tzoukermann-miller-2018-evaluating-error}
Evelyne Tzoukermann and Corey Miller. 2018.
\newblock \href {https://www.aclweb.org/anthology/W18-1922} {Evaluating
  automatic speech recognition in translation}.
\newblock In \emph{Proceedings of the 13th Conference of the Association for
  Machine Translation in the {A}mericas (Volume 2: User Track)}, pages
  294--302, Boston, MA. Association for Machine Translation in the Americas.

\bibitem[{Vaswani et~al.(2017)Vaswani, Shazeer, Parmar, Uszkoreit, Jones,
  Gomez, Kaiser, and Polosukhin}]{vaswani2017attention}
Ashish Vaswani, Noam Shazeer, Niki Parmar, Jakob Uszkoreit, Llion Jones,
  Aidan~N Gomez, {\L}ukasz Kaiser, and Illia Polosukhin. 2017.
\newblock \href
  {https://papers.nips.cc/paper/2017/file/3f5ee243547dee91fbd053c1c4a845aa-Paper.pdf}
  {Attention is all you need}.
\newblock In \emph{Advances in neural information processing systems}, pages
  5998--6008.

\bibitem[{Vijayakumar et~al.(2018)Vijayakumar, Cogswell, Selvaraju, Sun, Lee,
  Crandall, and Batra}]{diverse_beam_search}
Ashwin~K. Vijayakumar, Michael Cogswell, Ramprasaath~R. Selvaraju, Qing Sun,
  Stefan Lee, David~J. Crandall, and Dhruv Batra. 2018.
\newblock \href
  {https://www.aaai.org/ocs/index.php/AAAI/AAAI18/paper/view/17329} {Diverse
  beam search for improved description of complex scenes}.
\newblock In \emph{Proceedings of the Thirty-Second {AAAI} Conference on
  Artificial Intelligence, (AAAI-18)}, pages 7371--7379. {AAAI} Press.

\bibitem[{Wang et~al.(2020{\natexlab{a}})Wang, Tang, Ma, Wu, Okhonko, and
  Pino}]{Wang2020fairseqSF}
Changhan Wang, Yun Tang, Xutai Ma, Anne Wu, Dmytro Okhonko, and Juan Pino.
  2020{\natexlab{a}}.
\newblock \href {https://www.aclweb.org/anthology/2020.aacl-demo.6} {Fairseq
  {S}2{T}: Fast speech-to-text modeling with fairseq}.
\newblock In \emph{Proceedings of the 1st Conference of the Asia-Pacific
  Chapter of the Association for Computational Linguistics (AACL): System
  Demonstrations}, pages 33--39. Association for Computational Linguistics.

\bibitem[{Wang et~al.(2020{\natexlab{b}})Wang, Wu, and Pino}]{wang2020covost}
Changhan Wang, Anne Wu, and Juan Pino. 2020{\natexlab{b}}.
\newblock \href {https://arxiv.org/abs/2007.10310} {{CoVoST} 2: A massively
  multilingual speech-to-text translation corpus}.
\newblock \emph{arXiv preprint arXiv:2007.10310}.

\bibitem[{Wang et~al.(2020{\natexlab{c}})Wang, Wu, Liu, Yang, and
  Zhou}]{Wang_Wu_Liu_Yang_Zhou_2020}
Chengyi Wang, Yu~Wu, Shujie Liu, Zhenglu Yang, and Ming Zhou.
  2020{\natexlab{c}}.
\newblock \href {https://doi.org/10.1609/aaai.v34i05.6452} {Bridging the gap
  between pre-training and fine-tuning for end-to-end speech translation}.
\newblock \emph{Proceedings of the AAAI Conference on Artificial Intelligence},
  34(05):9161--9168.

\bibitem[{Watanabe et~al.(2018)Watanabe, Hori, Karita, Hayashi, Nishitoba,
  Unno, {Enrique Yalta Soplin}, Heymann, Wiesner, Chen, Renduchintala, and
  Ochiai}]{watanabe2018espnet}
Shinji Watanabe, Takaaki Hori, Shigeki Karita, Tomoki Hayashi, Jiro Nishitoba,
  Yuya Unno, Nelson {Enrique Yalta Soplin}, Jahn Heymann, Matthew Wiesner,
  Nanxin Chen, Adithya Renduchintala, and Tsubasa Ochiai. 2018.
\newblock \href {https://doi.org/10.21437/Interspeech.2018-1456} {{ESPnet}:
  End-to-end speech processing toolkit}.
\newblock In \emph{Proc. Interspeech 2018}, pages 2207--2211.

\bibitem[{Watanabe et~al.(2017)Watanabe, Hori, Kim, Hershey, and
  Hayashi}]{watanabe2017}
Shinji Watanabe, Takaaki Hori, Suyoun Kim, John~R. Hershey, and Tomoki Hayashi.
  2017.
\newblock \href {https://doi.org/10.1109/JSTSP.2017.2763455} {Hybrid
  {CTC}/attention architecture for end-to-end speech recognition}.
\newblock \emph{{IEEE} Journal of Selected Topics in Signal Processing},
  11(8):1240--1253.

\bibitem[{Weiss et~al.(2017)Weiss, Chorowski, Jaitly, Wu, and Chen}]{weiss2017}
Ron~J. Weiss, Jan Chorowski, Navdeep Jaitly, Yonghui Wu, and Zhifeng Chen.
  2017.
\newblock \href {https://doi.org/10.21437/Interspeech.2017-503}
  {Sequence-to-sequence models can directly translate foreign speech}.
\newblock In \emph{Proc. Interspeech 2017}, pages 2625--2629.

\bibitem[{Wiseman and Rush(2016)}]{wiseman2016sequence}
Sam Wiseman and Alexander~M. Rush. 2016.
\newblock \href {https://doi.org/10.18653/v1/D16-1137} {Sequence-to-sequence
  learning as beam-search optimization}.
\newblock In \emph{Proceedings of the 2016 Conference on Empirical Methods in
  Natural Language Processing}, pages 1296--1306, Austin, Texas. Association
  for Computational Linguistics.

\bibitem[{Yang et~al.(2018)Yang, Huang, and Ma}]{yang2018breaking}
Yilin Yang, Liang Huang, and Mingbo Ma. 2018.
\newblock \href {https://doi.org/10.18653/v1/D18-1342} {Breaking the beam
  search curse: A study of (re-)scoring methods and stopping criteria for
  neural machine translation}.
\newblock In \emph{Proceedings of the 2018 Conference on Empirical Methods in
  Natural Language Processing}, pages 3054--3059, Brussels, Belgium.
  Association for Computational Linguistics.

\bibitem[{Zhao et~al.(2020)Zhao, Wang, and Li}]{zhao2020neurst}
Chengqi Zhao, Mingxuan Wang, and Lei Li. 2020.
\newblock \href {https://arxiv.org/pdf/2012.10018.pdf} {{NeurST}: Neural speech
  translation toolkit}.
\newblock \emph{arXiv preprint arXiv:2012.10018}.

\bibitem[{Zheng et~al.(2021)Zheng, Chen, Ma, and Huang}]{zheng2021fused}
Renjie Zheng, Junkun Chen, Mingbo Ma, and Liang Huang. 2021.
\newblock \href {https://arxiv.org/pdf/2102.05766.pdf} {Fused acoustic and text
  encoding for multimodal bilingual pretraining and speech translation}.
\newblock \emph{arXiv preprint arXiv:2102.05766}.

\end{thebibliography}
\bibliographystyle{acl_natbib}

\newpage
\appendix
\section{Appendix}
\definecolor{cfred}{HTML}{e9a3c9}
\definecolor{cfgreen}{HTML}{a1d76a}

\begin{table*}[t]
  \centering
    
\resizebox {\linewidth} {!} {
\begin{tabular}{l|l|l}
\toprule
Model / Source & \multicolumn{1}{c|}{ASR Output} & \multicolumn{1}{c}{ST Output} \\
\midrule
Ground-Truth & \ldots~porque tengo \colorbox{cfgreen}{a mis dos hijos} acá & \ldots~because i have \colorbox{cfgreen}{my two children} here \\
Multi-Decoder & \ldots~porque tengo \colorbox{cfred}{mis dos hijos} acá & \ldots~because i have \colorbox{cfred}{two kids} here \\
\hspace{0.5em} +Speech-Attention & \ldots~porque tengo \colorbox{cfred}{mis dos hijos} acá & \ldots~because i have \colorbox{cfgreen}{my two children} here \\
\midrule
Ground-Truth & puedes ayudar para que \colorbox{cfgreen}{se haga justicia} más rápido  & you can help \colorbox{cfgreen}{so that justice} is served quickly \\
Multi-Decoder & puedes ayudar para que \colorbox{cfred}{sea justicia} más rápido & you can help \colorbox{cfred}{so it's} faster \\
\hspace{0.5em} +Speech-Attention & puedes ayudar para que \colorbox{cfred}{sea justicia} más rápido & you can help \colorbox{cfgreen}{so that it's} faster \colorbox{cfgreen}{justice} \\
\midrule
Ground-Truth & pero \colorbox{cfgreen}{tiene} muchas cosas muy bonitas & but \colorbox{cfgreen}{there are} many beautiful things \\
Multi-Decoder & pero \colorbox{cfred}{tienen} muchas cosas muy bonitas & but \colorbox{cfred}{they have} a lot of nice things \\
\hspace{0.5em} +Speech-Attention & pero \colorbox{cfred}{tienen} muchas cosas muy bonitas & but \colorbox{cfgreen}{there are} many very beautiful things \\
\midrule
Ground-Truth & \colorbox{cfgreen}{acampar} ir a pescar y ir a las monta\~nas a esquiar & \colorbox{cfgreen}{camping} and fishing and going to the mountains to ski \\
Multi-Decoder & \colorbox{cfgreen}{acampar} y a pescar y y de las monta\~nas esquiar & \colorbox{cfgreen}{camping} and fishing and and the mountains skiing \\
\hspace{0.5em} +Speech-Attention & \colorbox{cfred}{a campar} y ir a pescar y ir a las monta\~nas a esquiar & \colorbox{cfgreen}{camping} and go fishing and go to the mountains to ski \\
\bottomrule
\end{tabular}
}
    \caption{Examples where the Multi-Decoder and Multi-Decoder w/ Speech-Attention models make errors in the ASR portion of Spanish-English ST. In these cases the Speech-Attention component alleviates ASR error propagation, producing correct translations despite mistakes in transcription. Words that are transcribed/translated correctly are highlighted in \colorbox{cfgreen}{green} and those that are incorrect are in \colorbox{cfred}{pink}. }
    \label{tab:qual_results}
\end{table*}
\begin{table*}[t]
  \centering
\begin{tabular}{lcccccc}
\toprule
& \multicolumn{3}{c}{Fisher test} & \multicolumn{3}{c}{CallHome test} \\
\cmidrule(r){2-4}\cmidrule(r){5-7}
Model & BLEU ($\uparrow$) & METEOR($\uparrow$) & TER($\downarrow$) & BLEU ($\uparrow$) & METEOR($\uparrow$) & TER($\downarrow$) \\
\midrule
Baseline Enc-Dec & 49.5 & 37.9 & 42.7 & 18.2 & 22.9 & 68.7 \\ \midrule
Multi-Decoder & 52.6 & 39.7 & 40.5 & 20.1 & 24.6 & 66.5 \\
\hspace{0.5em}+ASR Re-scoring & 53.7 & 40.0 & 39.6 & 20.8 & 24.9 & 65.3 \\
\hspace{0.5em}+Speech-Attention & 54.1 & 40.2 & 39.2 & 21.4 & 25.2 & 65.3 \\
\hspace{1em}+ASR Re-scoring & \textbf{55.0} &  \textbf{40.4} & \textbf{38.5} & \textbf{21.5} & \textbf{25.4} & \textbf{64.2} \\
\bottomrule
\end{tabular}
    \caption{Results presenting the performance of our Baseline Enc-Dec implementation and our Multi-Decoder models as evaluated by three metrics: BLEU, METEOR, and Translation Edit Rate (TER). These are the same models as in \Tref{tab:main_results}, which uses BLEU. All results are from the Fisher-CallHome Spanish-English test corpus.}
    \label{tab:metrics_results}
\end{table*}

\subsection{Training and Inference hyperparameters}
\label{sec:appendix}
We tune training and inference hyperparameters using only the dev sets. We first determined the best hyperparameters for our baseline Enc-Dec implementation and fixed all settings not pertaining to the unique searchable hidden intermediates of our Multi-Decoder. Then, we find the best hyperparameters for our proposed models under these constraints to demonstrate a true comparison against the baseline. For our Speech-Attention variant, we found that increasing attention dropout in the ST sub-net decoder to 0.4 improved performance, which we verified was not true for the vanilla Multi-Decoder model. For our external model re-scoring, we found that a CTC weight of 0.3 is best for all Multi-Decoder and Multi-Decoder w/ Speech-Attention. The best LM weight for the Multi-Decoder was 0.2, while the best LM weight for the Multi-Decoder w/ Speech-Attention was 0.4. For both of these re-scoring hyperparameters, we tried $[0.2, 0.3, 0.4]$. For deciding the beam size, we use the experiment demonstrated in \Fref{fig:beam_exp} which uses beam sizes of $[1, 4, 8, 10, 16]$.

\subsection{Multi-Decoder ST Performance across other automatic MT Metrics}
\label{sec:other_metrics}
To supplement our overall ST results on the Fisher/CallHome corpus in \Tref{tab:main_results}, which shows BLEU scores, we also evaluated the same Multi-Decoder and Baseline Enc-Dec (Our Implementation) models on two additional metrics: METEOR \cite{banerjee2005meteor} and Translation Edit Rate (TER) \cite{snover2006study}. Performance across all three metrics show consistent trends, with the Multi-Decoder outperforming the Baseline Enc-Dec model on all metrics. We see that both the Multi-Decoder and Multi-Decoder w/ Speech-Attention models are improved through ASR Re-scoring. Further, the models with Speech-Attention perform better than those without.

\subsection{Qualitative Examples of Error Propagation Avoidance}
\label{sec:qual_examples}
To supplement our qualitative analysis of the error propagation avoidance of the Multi-Decoder with Speech-Attention model in \Sref{sec:error_prop}, we also show four qualitative examples in \Tref{tab:qual_results}. In the first three examples, the Multi-Decoder and Multi-Decoder with Speech-Attention models both make the same mistakes in the ASR portion of Spanish-English translation, but the model with Speech-Attention recovers by producing correct English translations despite mistakes in the Spanish transcription. On the other hand, the model without Speech-Attention propagates the Spanish transcription errors into English translation errors. In the fourth example only the Multi-Decoder w/ Speech-Attention makes a mistake in Spanish transcription, but the English translation still recovers.

\subsection{MuST-C Data Setup and Model Details}
\label{sec:mustc_data}
\paragraph{Data:} We extend our approach to other language pairs from the MuST-C speech translation corpus \cite{di2019must}. These are recordings of TED talks in English with translations in various target languages. In our experiments we show results on two language pairs, namely, English-German and English-French. We use the provided dev set for deciding the training and inference hyperparameters, as mentioned in Appendix (\ref{sec:appendix}). We report detokenized case-sensitive BLEU \cite{post-2018-call} on the tst-COMMON set. We apply the same text processing as done in \cite{inaguma-etal-2020-espnet-st} and use a joint source and target vocabulary of 8K byte pair encoding (BPE) units \cite{kudo2018sentencepiece}. Similar to \Sref{sec:data_prep}, we use the ESPnet library to prepare the corpus, and apply the same data preparation and augmentations.

\paragraph{Multi-Decoder Configuration:} For the MuST-C experiments, we scaled our Multi-Decoder w/ Speech-Attention config from the Fisher-CallHome experiments by increasing the $\textsc{Encoder}_{\textsc{st}}$ to contain 4 transformer encoder blocks. We increased the attention dim and attention heads of the $\textsc{Encoder}_{\textsc{asr}}$ and $\textsc{Decoder}_{\textsc{asr}}$ to 512 dimension and 8 heads respectively, while only increasing the attention dimension to 512 for $\textsc{Encoder}_{\textsc{st}}$ and $\textsc{Decoder}_{\textsc{st}}$. This increased the total trainable parameters to 135M, which we trained on 4 NVIDIA V-100 GPUs for $\approx$3 days. We also found that increasing the attention dropout of ASR decoder to 0.2 helped with the increased parameters. We kept the remaining dropout parameters the same as our previous experiments. We also keep the remaining training configurations the same like the effective batch-size, learning rate and warmup steps, loss weighting and SpecAugment policy.

During inference, we use the same beam sizes from our Fisher-CallHome experiments and we perform a search across the length penalty and max length ratio settings using the MuST-C dev sets.
In the intermediate ASR beam search we use a length penalty of 0.1 and 0.2 for English-German and English-French respectively. In the ST beam search we use a max length ratio of 0.3 and length penalties of 0.6 and 0.5 for English-German and English-French respectively. For our experiments with ASR re-scoring, we use a LM weight of 0.1 and a CTC weight of 0.1. In these re-scoring experiments we also set the ASR length penalty to 0.6 and the ST length penalty to 0.5, while increasing the ST max length ratio to 0.5. 
The LMs used were trained on the English transcripts of the MuST-C English-German and English-French corpora, with dev perplexities of 32.7 and 23.2 respectively.

\end{document}